\pdfoutput=1

\documentclass[11pt]{article}

\usepackage[final]{acl}
\usepackage{times}
\usepackage{latexsym}
\usepackage[T1]{fontenc}
\usepackage[utf8]{inputenc}
\usepackage{microtype}
\usepackage{inconsolata}
\usepackage{graphicx}
\usepackage{amssymb}
\usepackage{amsmath}
\usepackage{booktabs}
\usepackage{makecell}
\usepackage{multirow}
\usepackage{url}
\usepackage{bm}
\usepackage[noend]{algpseudocode}
\usepackage{algorithmicx,algorithm}
\usepackage{xcolor}
\usepackage{colortbl, booktabs}
\definecolor{mygreen}{HTML}{3C8205}
\definecolor{myred}{HTML}{80032E}
\definecolor{mypurple}{rgb}{0.925, 0.902, 0.969}
\definecolor{mygrey}{gray}{0.93}
\usepackage{listings}
\usepackage{hyperref}
\usepackage{fontawesome5}
\lstdefinestyle{SQLStyle}{
  language=SQL,
  backgroundcolor=\color[HTML]{F7F7F7},   
  basicstyle=\ttfamily\small,          
  keywordstyle=\color{blue}\bfseries,  
  commentstyle=\color{green},          
  stringstyle=\color{red},             
  numbers=right,                        
  numberstyle=\tiny\color{gray},       
  stepnumber=1,                        
  numbersep=5pt,                       
  showspaces=false,                    
  showstringspaces=false,              
  breaklines=true,                     
  breakatwhitespace=true               
}

\makeatletter

\newcommand{\Rmnum}[1]{\expandafter\@slowromancap\romannumeral #1@}
\makeatother

\makeatletter
\renewcommand{\maketag@@@}[1]{\hbox{\m@th\normalsize\normalfont#1}}%
\makeatother

\title{Training a Utility-based Retriever Through Shared Context Attribution for Retrieval-Augmented Language Models}

\author{Yilong Xu$^{1,2,3}$\quad Jinhua Gao$^{1,2}$\thanks{Corresponding author.} \quad Xiaoming Yu$^{1,2}$\quad Yuanhai Xue$^{1,2}$ \quad Baolong Bi$^{1,2,3}$ \\ \textbf{Huawei Shen}$^{1,2,3}$\quad \textbf{Xueqi Cheng}$^{1,2,3}$\quad \\
$^1$State Key Lab of AI Safety, Institute of Computing Technology, CAS \\
$^2$Key Lab of AI Safety, Chinese Academy of Sciences \\
$^3$University of Chinese Academy of Sciences \\
\texttt{\small{\{xuyilong23s, gaojinhua\}@ict.ac.cn}}
}

\begin{document}

\maketitle

\begin{abstract}

Retrieval-Augmented Language Models boost task performance, owing to the retriever that provides external knowledge.
Although crucial, the retriever primarily focuses on semantics relevance, which may not always be effective for generation.
Thus, utility-based retrieval has emerged as a promising topic, prioritizing passages that provide valid benefits for downstream tasks. 
However, due to insufficient understanding, capturing passage utility accurately remains unexplored.
This work proposes SCARLet, a framework for training utility-based retrievers in RALMs, which incorporates two key factors, multi-task generalization and inter-passage interaction.
First, SCARLet constructs shared context on which training data for various tasks is synthesized.
This mitigates semantic bias from context differences, allowing retrievers to focus on learning task-specific utility and generalize across tasks.
Next, SCARLet uses a perturbation-based attribution method to estimate passage-level utility for shared context, which reflects interactions between passages and provides more accurate feedback.
We evaluate our approach on ten datasets across various tasks, both in-domain and out-of-domain, showing that retrievers trained by SCARLet consistently improve the overall performance of RALMs.

\noindent\textbf{Resources:} \faGithub\ \href{https://github.com/ylXuu/SCARLet}{github.com/ylXuu/SCARLet}

\end{abstract}

\section{Introduction}

Retrieval-Augmented Language Models \citep[RALMs;][]{rag_lewis_2020} typically comprise two parts: the retriever and the generator. The retriever collects up-to-date task-related external information, while the generator incorporates the collected non-parametric knowledge into inference. RALMs have achieved enhanced performance across various downstream tasks, including question answering, fact checking, and dialogue generation~\citep{shao2023enhancingretrievalaugmentedlargelanguage, lift_yourself_up}. As a crucial role, the optimization of the retrievers in RALMs has become a trending research topic.

Early RALMs adopt relevance-based retrievers, including both sparse~\citep{bm25} and dense~\citep{dpr} models. However, these retrievers are primarily biased toward semantic relevance~\citep{wu2024easilyirrelevantinputsskew}, failing to consider the passage utility and leading to misalignment in RALMs. The utility, measuring the valid gain that a passage contributes to the downstream generation~\citep{arellmgoodatutilityjudgements}, can bridge the gap between the retriever and the generator. Some recent works have proposed to optimize retrievers by constructing feedback from generators~\citep{replug, aar, wei2024instructrag}, achieving promising results. Nonetheless, how to align retrievers to better capture utility remains an open yet challenging problem.

Different from relevance, which is mainly determined by the query and the passage \citep{learningmaximalmarginal, matchzoo}, utility needs a more comprehensive measurement. In this paper, we propose the following two vital yet overlooked factors for utility modeling in RALMs:

\paragraph{Multi-task Generalization} RALMs need to accommodate various downstream tasks, where the utility of a passage can vary accordingly. Existing methods typically optimize retrievers using the pooling strategy, i.e., mixing data from different tasks for training, to learn task-specific retrieval criteria~\citep{lin2024raditretrievalaugmenteddualinstruction, stochasticrag}. However, since pooled samples from different tasks typically have different contexts, the trained retrievers might tend to capture semantic relevance signals instead of utility features. Such unexpected preference will downgrade the retrievers' generalization ability, especially for those with weaker linguistic capabilities~\citep{liu2024finegrainedguidanceretrieversleveraging}.

\paragraph{Inter-passage Interaction} In some complex tasks, the utility of a certain passage cannot be solely determined by itself. For example, when handling multi-hop question-answering tasks, the model should rely on preceding and even succeeding contexts in the reasoning chain to judge a passage's utility. However, the utility signals constructed in previous works either fail to provide passage-level feedback~\citep{stochasticrag, sohn2024rationaleguidedretrievalaugmentedgeneration} or evaluate each passage independently~\citep{aar, replug}, leading to imprecise utility measurements.

In this paper, we propose a novel framework to train utility-based retrievers for RALMs, named \textbf{SCARLet}, representing \underline{s}hared \underline{c}ontext \underline{a}ttribution supe\underline{r}vised training for uti\underline{l}ity-based r\underline{et}rievers.

Specifically, SCARLet first introduces a training data synthesis pipeline. Contrary to the previous pooling strategy that mixes training data with different contexts, our pipeline first constructs a shared context, and subsequently synthesizes training data for various downstream tasks derived from the shared context. This method mitigates the semantic interference by achieving single-variable control, and enables the retriever to focus on learning task-specific utility. To better assess the utility of certain passages, SCARLet employs a passage-level perturbation-based technique, which randomly removes some passages from the context and measures the fluctuations in the generated output. Such a design can effectively capture the synergy between passages, thereby accurately reflecting their utility. Finally, SCARLet collects positive and negative samples based on the utility scores and trains the retriever in a contrastive way.

We conduct extensive experiments to evaluate the performance gain brought by SCARLet. Our experiments adopt ten datasets, covering eight distinct tasks that are frequently used for RALMs evaluation. The results show that RALMs equipped with retrievers trained by SCARLet, consistently achieve optimal or suboptimal downstream performance across all datasets. Moreover, further analysis and case studies demonstrate that SCARLet can better capture utility signals.

To summarize, our main contributions include:

\begin{itemize}
    \item We argue that utility should be preferred in RALMs and propose two critical factors for training utility-based retrievers.
    \item We propose SCARLet, a novel framework to train utility-based retrievers through shared context synthesis and utility attribution.
    \item We conduct extensive experiments across various tasks, demonstrating that our proposed SCARLet can improve the overall performance of RALMs.
\end{itemize}

\section{Related Work}

\paragraph{RALMs} Large Language Models \citep[LLMs;][]{llm} exhibit remarkable performance across a wide range of tasks ~\citep{llmsurvey1, llmsurvey2, wei2022emergentabilitieslargelanguage}. However, LLMs also face the challenge of hallucinations, often performing poorly when addressing factual issues ~\citep{hallucination1, bi2024adaptivetokenbiaserknowledge}. The emergence of RALMs effectively alleviates the weakness of insufficient factuality ~\citep{ragsurvey}. A RALM system typically comprise a retriever and a generator, where the retriever recalls external information to enhance the generator to respond more accurately. To further optimize RALMs and improve the synergy between the two parts, existing methods generally fall into three categories: 1) overall optimization ~\citep{lin2024raditretrievalaugmenteddualinstruction, stochasticrag}; 2) generator-only optimization ~\citep{raat, rankrag, bi2025parameters}; 3) retriever-only optimization ~\citep{replug, aar}. Optimizing only the retriever is a more efficient and cost-effective way that offers plug-and-play capabilities, enhancing the overall efficiency and stability of the RALM system.

\begin{figure*}[htbp]
  \includegraphics[width=\linewidth,scale=1.00]{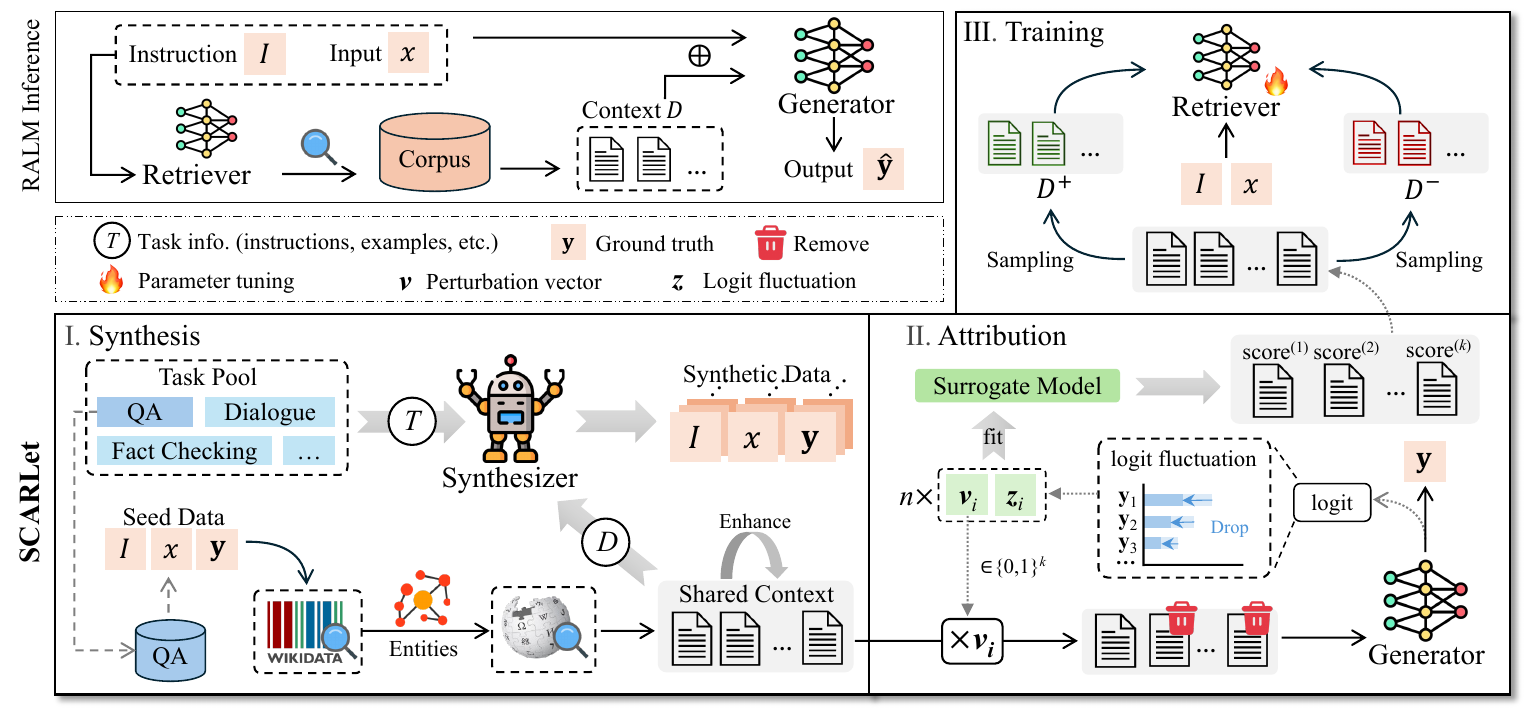}
  \caption {The illustration of SCARLet. The upper left part describes the inference process of RALMs. In SCARLet, there are three main stages. First, the shared context is constructed by retrieving external corpus based on the seed data. The synthesizer is instructed with shared context and task information from the task pool, to generate synthetic data. Next, using the shared context as the data source, SCARLet applies perturbation-based utility attribution on the generator, and then, based on the utility scores, samples positive and negative passages for retriever training.}
  \label{fig:fig-1}
\end{figure*}

\paragraph{Utility-based Retrieval} In RALMs, early exploration of retrieval utility focuses on capturing the downstream feedback of generators. \citet{towardsasearchengineformachines} propose supervision based on downstream task metrics, but fail to provide passage-level utility feedback. \citet{replug, aar} assess utility of each passage using generator outputs, but they ignore the interactions between passages. \citet{sohn2024rationaleguidedretrievalaugmentedgeneration, wei2024instructrag} employ the generator's self-reflection to evaluate utility, which may bring hallucinations as the language models can be dishonest ~\citep{madsen-etal-2024-self}. \citet{tart, glass-etal-2022-re2g} notice the multi-task nature of the retrieval stage, but fail to account for the training biases introduced by contextual differences in the pooling strategy.

Therefore, our proposed SCARLet framework comprehensively considers the above issues of multi-task generalization and utility assessment, offering a novel pipeline with shared context synthesis and utility attribution to effectively train utility-based retrievers in RALMs.

\section{Method}

In this section, we first define the RALMs system, then we introduce the SCARLet pipeline.

\subsection{Definitions}
\label{sec:3-1}

A typical RALM system consists of a retriever and a generator. During the retrieval stage, we employ a dense retriever based on an encoder $\mathbf{Enc}$ with parameters $\phi$. And the retriever interacts with an external corpus $\mathcal{C}$. For a query $q$, we calculate the dot product of the embeddings of $q$ and each passage $d$ in $\mathcal{C}$, as follows:
\begin{equation}
    \mathrm{score}\left ( q,d \right ) = \mathbf{Enc}_{\phi} \left ( q \right ) \cdot \mathbf{Enc}_{\phi} \left ( d \right ), d \in \mathcal{C}.
\end{equation}
The top-$k$ passages with the highest scores are selected and added to the context, denoted as $D = \left [d_1,\dots,d_k \right]$. Note that RALMs need to accommodate various downstream tasks, for a task $T$ and an input $x$ from its dataset, we define the query format as $q=I \oplus x$, where $I$ denotes the instruction description of task $T$. 

In the generation stage, a language model $\mathrm{LM}$ with parameters $\theta$ serves as the generator. The context $D$ is used to enhance generation, ultimately producing the predicted output $\mathbf{\hat{y}}$, as shown below:
\begin{equation}
    \mathbf{\hat{y}} =\mathrm{LM}_{\theta }\left (I \oplus x \oplus D \right ) ,
\end{equation}
where $\mathbf{\hat{y}}$ is a sequence and $\mathbf{\hat{y}}_t$ denotes the $t$-th token. We denote the ground truth of $x$ as $\mathbf{y}$.

\subsection{Overview of SCARLet}

The overall architecture of our proposed SCARLet is shown in Figure \ref{fig:fig-1}, including shared context synthesis and training data construction (\S \ref{sec:3-2}), utility attribution modeling (\S \ref{sec:3-3}), as well as data sampling and retriever tuning (\S \ref{sec:3-4}).

Shared context refers to the common context for data of different tasks in the training stage, which is then used to enhance downstream generation. Previous studies employ the pooling strategy~\citep{lin2024raditretrievalaugmenteddualinstruction, stochasticrag}, where each instance has a distinct context for training. Learning task-specific features to improve multi-task generalization of utility-based retrieval might be disturbed by the semantically relevant noise introduced by differences in context, leading to unexpected preference, particularly in retrievers with weaker linguistic capabilities. To tackle the above challenges, our proposed SCARLet adopts a reverse strategy, first constructing shared context to narrow the semantic gap, and then synthesizing task-specific data based on this context. Sharing context across tasks can highlight utility feature differences, making it easier to learn. Moreover, LLM-driven data synthesis has been shown to be a promising way \citep{long2024llmsdrivensyntheticdatageneration,kim2025syntrievertrainretrieversynthetic}, which can effectively reduce labor costs.

Utility attribution modeling refers to local explanation techniques to build utility signals from the downstream generation. More specifically, we adopt the contributive attribution model, which measures how the input context contributes to the model's output and aligns well with the definition of utility in RALMs. Previous research on optimizing retrievers from downstream generation, either fails to construct passage-level feedback or only considers the individual impact of each passage, overlooking the synergy effects between passages. Therefore, taking the shared context as the source data, SCARLet uses a passage-level perturbation-based utility attribution approach, where fluctuations in generation caused by perturbations can reflect interactions between passages and then be quantified as utility scores.

\subsection{Shared Context Synthesis}
\label{sec:3-2}

Specifically, we first define a task pool $\mathcal{T}$, which is linked to various downstream tasks and their datasets, such as multi-hop QA, long-form QA, and fact checking. We begin by collecting seed data from datasets of $\mathcal{T}$, including task instructions, inputs, and ground truth. In line with the motivation behind shared context, passages within this context need to be closely related to facilitate the synthesis of high-quality data. Therefore, we employ an approach based on associated entities, which extracts entities from the seed data, searches their adjacent entities by querying Wikidata\footnote{\url{https://www.wikidata.org}}, and merges them to obtain a related entity list. We then use this list to retrieve relevant passages from the Wikipedia corpus, and treat the recalled passages as the shared context $D_{\mathrm{shared}}$. Subsequently, we instruct the synthesizer model $\mathsf{S}$ to generate new training data, using $D_{\mathrm{shared}}$ as the information source and task information (including instructions and examples) from $\mathcal{T}$. The process is formalized as follows:
\begin{equation}
    \left ( x^{\mathrm{new} }_{T_1}, \mathbf{y}^{\mathrm{new} }_{T_1} \right ),\dots,\left ( x^{\mathrm{new} }_{T_{l}},\mathbf{y}^{\mathrm{new} }_{T_{l}} \right )=\mathsf{S}\left ( D_{\mathrm{shared} },\mathcal{T} \right ),
\end{equation}
where $x^{\mathrm{new}}_{T_i}$ and $\mathbf{y}^{\mathrm{new}}_{T_i}$ represent input and ground truth of the synthetic data of task $T_i$, respectively. $l$ is the total number of tasks in $\mathcal{T}$.

To improve the quality of synthetic data, the task pool $\mathcal{T}$ not only provides the task instructions but also offers example data. To further improve robustness, following \citet{raat, arellmgoodatutilityjudgements}, we also introduce synthetic noise into the shared context by instructing the synthesizer to generate semantically relevant but useless passages. In addition, we incorporate a data filtering step that instructs the synthesizer to eliminate samples containing faults. For more details, please refer to Appendix \ref{sec:appendix_a}. We also provide an example of shared context in Appendix \ref{sec:appendix_f}.

\subsection{Passage-level Utility Attribution}
\label{sec:3-3}

Specifically, the context $D$ recalled by the upstream retriever consists of $k$ passages. To evaluate the utility of each individual passage with inter-passage interactions, we adopt a perturbation-based method where we remove certain passages and inspect the changes in the final output. The approach is implemented via introducing a perturbation vector $\mathbf{v} \in \left \{ 0,1 \right \}^{k}$, where 0 and 1 indicate whether the corresponding passage is removed or included, respectively. However, running all generations of $2^{k}$ possible perturbation vectors can result in significant computational overhead. Inspired by the method of Local Interpretable Model-agnostic Explanations \citep[LIME;][]{lime, ananalysisoflimefortextdata}, we first sample $n$ perturbation vectors randomly and then fit a surrogate model for predicting the utility score, as shown below:
\begin{equation}
    \hat{\bm{\alpha} }  \in \mathop{\arg \min}_{\bm{\alpha} \in \mathbb{R}^{k + 1} } \left \{ \sum_{i=1}^{n} \left ( z_i-\bm{\alpha}^T \mathbf{v}_i  \right )^2 + \lambda \left \| \bm{\alpha}  \right \|^2  \right \},
\end{equation}
where we adopt the ridge regression~\citep{Hilt1977RidgeAC} as our surrogate model, $\bm{\alpha}$ represents the parameters to be fitted, $\lambda$ is a hyperparameter for regularization, and $z_i$ is the observed value under $\mathbf{v}_i$. More specifically, $\bm{\alpha}^{(i)}$ denotes the utility score of passage $d_i$, $\bm{\alpha}^{(0)}$ represents the intercept term. And $z_i$, which quantifies the fluctuation caused by $\mathbf{v}_i$, is calculated using the logit values of the tokens in the ground truth $\mathbf{y}$ at each time step, as shown below:
\begin{equation}
    z_i =  \sum_{t} \mathrm{logit} \left ( \mathbf{y}^{\left ( i \right ) }_t \right ).
\end{equation}

\begin{figure}[t!]
  \centering
  \includegraphics[width=0.8\linewidth]{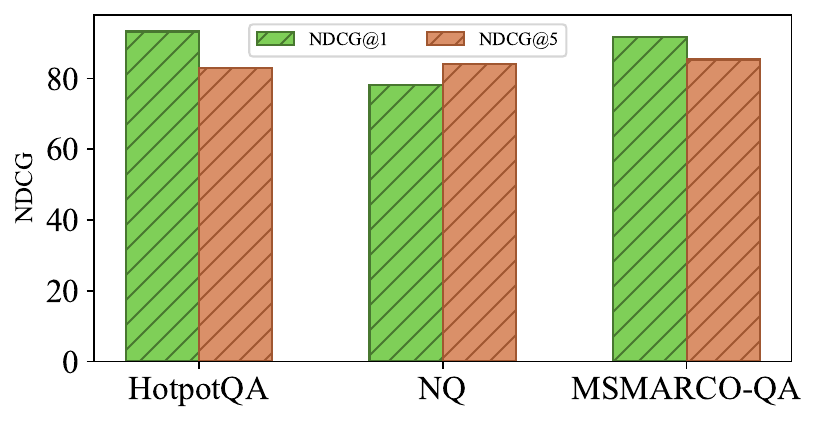}
  \caption {The performance of the perturbation-based attribution method on the GTI benchmark. The nDCG metrics show that it achieves at least about 80\% performance on three datasets, with some exceeding 90\%.}
  \label{fig:fig-2}
\end{figure}

To evaluate the effectiveness of the above utility attribution method, we conduct a preliminary experiment on the GTI benchmark~\citep{arellmgoodatutilityjudgements}, which includes three datasets: HotpotQA ~\citep{yang-etal-2018-hotpotqa}, Natural Questions ~\citep[NQ;][]{kwiatkowski-etal-2019-nq}, and MSMARCO-QA ~\citep{msmarcoqa}. Each test sample comprises input, ground truth, and ten passages including correct passages and other noise passages. We use the utility score to rank the passages. The results, measured using nDCG, demonstrate that our method shows a high accuracy in reflecting passage utility, as shown in Figure \ref{fig:fig-2}. We also compare our method to other attribution approaches, and our method outperforms them by over 20\%. For further details of the experiment, please refer to Appendix \ref{sec:appendix_b}.

\subsection{Sampling and Training}
\label{sec:3-4}

\begin{figure}[t!]
  \centering
  \includegraphics[width=\linewidth]{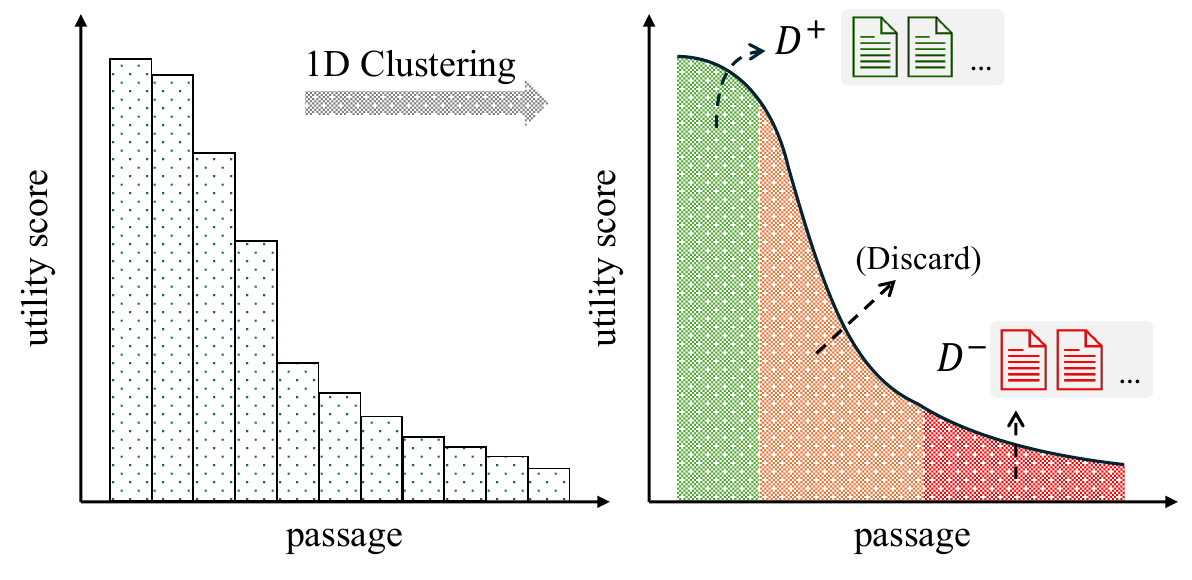}
  \caption {The illustration of the 1D clustering sampling. Based on the utility score, this method clusters the passages into three labels: the high-score passages (green) corresponding to positive samples, the middle-score passages (orange) that will be discarded, and the low-score passages (red) corresponding to negative samples.}
  \label{fig:fig-3}
\end{figure}

After calculating the utility score for each passage in the shared context, we then collect positive and negative samples based on these scores for training the retriever. When sorted in descending order of the scores, the utility distribution follows an inverse S-shaped curve, as depicted in Figure \ref{fig:fig-3}. Passages with higher scores correspond to positive samples, while those with lower scores represent negative samples. To effectively sample these two types of data, we employ a one-dimensional clustering approach. Specifically, we take the utility score list as the input and divide it into three clusters: one for the positive samples, one for the intermediate samples that will be discarded, and another for the negative samples. This method can dynamically adjust the number of useful passages in the context on various tasks and data.

After obtaining positive and negative samples, following \citet{ance}, the loss function is calculated as follows:
\begin{multline}
    \mathcal{L} = \sum_{x}  \sum_{d^+ \in D^+} \sum_{d^- \in D^-} \\
    l \left ( \mathrm{score} \left ( x,d^+ \right ), \mathrm{score} \left ( x,d^- \right )   \right ),
\end{multline}
where $l$ represents the cross-entropy loss.

\section{Experimental Setup}

This section introduces the main experiment setup, including datasets, baselines and implementation.

\begin{table}[t!]
\centering
\small
\resizebox{\linewidth}{!}{
\renewcommand\arraystretch{1.3}
\begin{tabular}{cccc}
    \toprule
    \textbf{Dataset} & \textbf{Task} & \textbf{Corpus} & \textbf{Metric} \\
    \midrule
    \rowcolor{gray!30} \multicolumn{4}{c}{\textbf{In-domain}} \\
    NQ~\citep{kwiatkowski-etal-2019-nq} & Single-hop QA & Wikipedia & Accuracy \\
    HotpotQA~\citep{yang-etal-2018-hotpotqa} & Multi-hop QA & Wikipedia & Accuracy \\
    ELI5~\citep{fan-etal-2019-eli5} & Long-form QA & Wikipedia & ROUGE-L \\
    FEVER~\citep{Thorne18Fever} & Fact checking & Wikipedia & Accuracy \\
    WoW~\citep{dinan2019wizard} & Dialogue generation & Wikipedia & F1 \\
    T-REx~\citep{elsahar-etal-2018-trex} & Slot filling & Wikipedia & Accuracy \\
    \rowcolor{gray!30} \multicolumn{4}{c}{\textbf{Out-of-domain}} \\
    zs-RE~\citep{levy-etal-2017-zero} & Relation extraction & Wikipedia & Accuracy \\
    SciFact~\citep{wadden-etal-2020-scifact} & Fact checking & BeIR & Accuracy \\
    Climate-FEVER~\citep{diggelmann2021climatefeverdatasetverificationrealworld} & Fact checking & BeIR & Accuracy \\
    FiQA~\citep{fiqa} & Financial QA & BeIR & ROUGE-L \\
    \bottomrule
\end{tabular}
}
\caption{The datasets used in the main experiment. Climate-Fever is a four-class classification task, while the other two fact-checking tasks are binary. For metrics, NQ, HotpotQA, T-REx, and zs-RE all calculate accuracy based on exact substring matching.}
\label{tab:tb-1}
\end{table}

\begin{table*}[t!]
\centering
\small
\resizebox{\linewidth}{!}{
\renewcommand\arraystretch{1.1}
\begin{tabular}{l|cccccc|cccc}
    \toprule
    \multirow{3}{*}{\textbf{Method}} & \multicolumn{6}{c}{\textbf{In-domain}} & \multicolumn{4}{c}{\textbf{Out-of-domain}} \\
    \cmidrule(lr){2-7} \cmidrule(lr){8-11}
     & \textbf{NQ} & \textbf{HotpotQA} & \textbf{ELI5} & \textbf{FEVER} & \textbf{WoW} & \textbf{T-REx} & \textbf{zs-RE} & \textbf{SciFact} & \textbf{C-FEVER} & \textbf{FiQA} \\
     
     \rowcolor{gray!30} \multicolumn{11}{c}{\textbf{LLaMA-3-8B-Instruct}} \\
     No retrieval & 43.5 & 36.8 & 14.8 & 79.8 & 9.3 & 34.5 & 21.7 & 68.0 & \underline{45.8} & 17.2 \\
     \midrule
     Contriever & 43.8 & 36.7 & 14.5 & 78.5 & 8.6 & 33.6 & 20.4 & 70.1 & 38.2 & 16.2 \\
     BGE & \underline{47.5} & \underline{41.6} & 15.2 & \textbf{83.5} & 8.7 & \underline{36.4} & 22.7 & \textbf{83.3} & 44.9 & \underline{21.0} \\
     $\textrm{AAR}_{\textrm{Contriever}}$ & 44.9 & 39.9 & 15.0 & 77.2 & 8.3 & 34.4 & 21.0 & 73.6 & 39.2 & 16.7 \\
     $\textsc{RePlug}_{\textrm{Contriever}}$ & 43.3 & 38.9 & 13.8 & 80.0 & 9.4 & 33.1 & \underline{22.8} & 74.6 & 41.2 & 18.9  \\
     \midrule
     \rowcolor{mypurple}
     $\textrm{SCARLet}_{\textrm{Contriever}}$ & 44.6 & 40.5 & \underline{15.8} & 80.6 & \underline{11.0} & 35.8 & 21.0 & 75.5 & 42.8 & 17.7  \\
     \rowcolor{mypurple}
     $\textrm{SCARLet}_{\textrm{BGE}}$ & \textbf{49.2} & \textbf{47.0} & \textbf{16.3} & \underline{81.3} & \textbf{12.2} & \textbf{37.0} & \textbf{24.4} & \underline{82.2} & \textbf{46.1} & \textbf{22.9} \\

     \rowcolor{gray!30} \multicolumn{11}{c}{\textbf{Qwen2.5-3B-Instruct}} \\
     No retrieval & 27.4 & 26.5 & \textbf{15.2} & 66.1 & 11.5 & 26.0 & 7.3 & 58.2 & \textbf{40.4} & 17.7 \\
     \midrule
     Contriever & 32.6 & 28.8 & 14.3 & 67.0 & 10.5 & 27.2 & 14.3 & 64.9 & 31.6 & 15.5 \\
     BGE & \textbf{46.8} & \underline{39.6} & 13.7 & \textbf{78.2} & 10.4 & \underline{29.3} & 15.5 & \textbf{70.6} & 30.2 & 18.7 \\
     $\textrm{AAR}_{\textrm{Contriever}}$ & 34.1 & 29.7 & 13.8 & 66.6 & 10.1 & 28.7 & 15.2 & 63.6 & 32.2 & 16.1 \\
     $\textsc{RePlug}_{\textrm{Contriever}}$ & 33.7 & 34.0 & 14.0 & 71.4 & \underline{12.2} & 26.9 & 16.2 & 61.1 & 30.6 & \underline{19.0} \\
     \midrule
     \rowcolor{mypurple}
     $\textrm{SCARLet}_{\textrm{Contriever}}$ & 38.2 & 35.4 & \underline{14.9} & 70.8 & 11.7 & 28.0 & \textbf{19.1} & \underline{65.3} & 31.7 & 17.3 \\
     \rowcolor{mypurple}
     $\textrm{SCARLet}_{\textrm{BGE}}$ & \underline{44.9} & \textbf{41.1} & \textbf{15.2} & \underline{74.3} & \textbf{12.6} & \textbf{29.7} & \underline{16.6} & 62.3 & \underline{33.0} & \textbf{20.4} \\

     \bottomrule
\end{tabular}
}
    \caption{Results of the main experiment across datasets on different downstream generators. $\textrm{AAR}_{\textrm{Contriever}}$, $\textsc{RePlug}_{\textrm{Contriever}}$, $\textrm{SCARLet}_{\textrm{Contriever}}$ represent the baselines initilized from Contriever, and $\textrm{SCARLet}_{\textrm{BGE}}$ represents the baseline initialized from BGE-base-v1.5. The \textbf{bold} score means the best performance of the corresponding dataset among baselines within the same generator, while the \underline{underline} score means the second best.}
    \label{tab:tb-2}
\end{table*}

\subsection{Datasets and Evaluation}
\label{sec:4-1}

We collect both in-domain and out-of-domain datasets for our experiments. In-domain datasets are utilized for providing seed data to construct synthetic training data, while out-of-domain datasets possess different tasks and corpora and are collected for further generalization tests. We collect seven datasets from KILT~\citep{petroni-etal-2021-kilt}, and three from BeIR~\citep{thakur2021beirheterogenousbenchmarkzeroshot}, as detailed in Table \ref{tab:tb-1}. All KILT datasets utilize Wikipedia dump dated 2019-08-01\footnote{\url{http://dl.fbaipublicfiles.com/BLINK/enwiki-pages-articles.xml.bz2}} as the corpus. Following \citet{wang-etal-2019-multi}, we split the original articles into segments with a maximum length of 100 words, resulting in a total of 28,773,800 passages. For test sets of BeIR, we adopt their self-constructed corpora. For retrieval, we follow the closed corpus setup\citep{tart}, where RALMs only retrieve from the corpus of the current dataset. For the test data, we randomly sample 1,000 data from the test split of each dataset.

For evaluation metrics, we mainly assess the performance of downstream tasks. For WoW, we use F1. For ELI5 and FiQA, we use ROUGE-L. For other datasets, we use accuracy.

\subsection{Baselines}
\label{sec:4-2}

The baselines are categorized into three settings:

\paragraph{No Retrieval} The downstream generators operate without any retrieval.

\paragraph{Vanilla RAG} Retrievers are added and the recalled passages are incorporated into the generation process. We choose two well-trained embedding models, Contriever~\citep{contriever} and BGE-base-v1.5~\citep{bge_embedding} as the retrievers.

\paragraph{Retriever-only Optimization} Retrievers are optimized using feedback from the generator. We select two recent methods, RePlug~\citep{replug} and AAR~\citep{aar}, both of which are initialized from Contriever.

We utilize LLaMA-3-8B-Instruct~\citep{llama3modelcard} and Qwen2.5-3B-Instruct~\citep{qwen2.5} as the generators in RALMs. All retrieval-based baselines use the top-3 passages. Given that some retrievers may not be tuned by instructions, the query format for Contriever and its baselines only contains $x$, without task instruction $I$. For BGE and its baselines, the query format follows the definition in Section \ref{sec:3-1}, which contains both $x$ and $I$.

\subsection{Implementation Details}
\label{sec:4-3}

In the shared context synthesis stage, we add the six tasks of the in-domain datasets into the task pool. We then randomly sample 1,000 data from the training split of each dataset to construct the seed dataset. We only consider one-hop relation when searching adjacent entities. For each entity, the top-10 passages are retrieved from $\mathcal{C}$, and the shared context is formed by selecting the top-10 passages across all retrieved passages. We utilize gpt-4o-2024-11-20~\citep{gpt4o} as the synthesizer model. For more implementation details and meta data, please see Appendix \ref{sec:appendix_c}.

\section{Results}

In this section, we present the results of main experiment (\S \ref{sec:5-1}), ablation study (\S \ref{sec:5-2}), retrieval evaluation (\S \ref{sec:5-3}), and case study (\S \ref{sec:5-4}).

\subsection{Overall Performance}
\label{sec:5-1}

The main experimental results are shown in Table \ref{tab:tb-2}. Our proposed SCARLet method achieves either optimal or suboptimal performance across various datasets and generators, demonstrating its effectiveness. Our detailed analysis from different perspectives is as follows:

\paragraph{In-domain Performance} In the evaluation on six in-domain datasets, the retrievers trained by SCARLet achieve the best performance in five datasets when using LLaMA-3-8B as the generator, and in four datasets when using Qwen-2.5-3B as the generator. Except for NQ and FEVER, SCARLet consistently outperforms the initial baselines, including Contriever and BGE.

\paragraph{Out-of-domain Performance} In the evaluation on four out-of-domain datasets, SCARLet also achieves optimal or suboptimal results. Specifically, SCARLet can still show progress in SciFact, Climate-FEVER, and FiQA, whose corpora differ from the Wikipedia corpus used in training and whose domains are notably different from the in-domain datasets, highlighting its generalization across corpora. In addition, SCARLet can achieve overall improvements when using two different downstream LLMs, preliminarily indicating its adaptability across generators.

\subsection{Ablation Study}
\label{sec:5-2}

According to the pipeline of SCARLet, we design the ablation experiments from three stages: 1) In the data synthesis stage, we evaluate the method of removing the step of retrieving adjacent entities, and instead directly retrieving the top-$k$ passages from the corpus $\mathcal{C}$ using only the entities extracted from the seed data; 2) In the utility attribution stage, since Section \ref{sec:3-3} already compares various attribution methods and demonstrates the superiority of our perturbation-based approach, we no longer conduct ablation study for this part; 3) In the sampling and training stage, we assess the effect of removing the one-dimensional clustering step and instead directly selecting the highest-scoring passage as the positive sample and the five lowest-scoring passages as negative samples based on the scores.

\begin{figure}[t!]
  \centering
  \includegraphics[width=\linewidth]{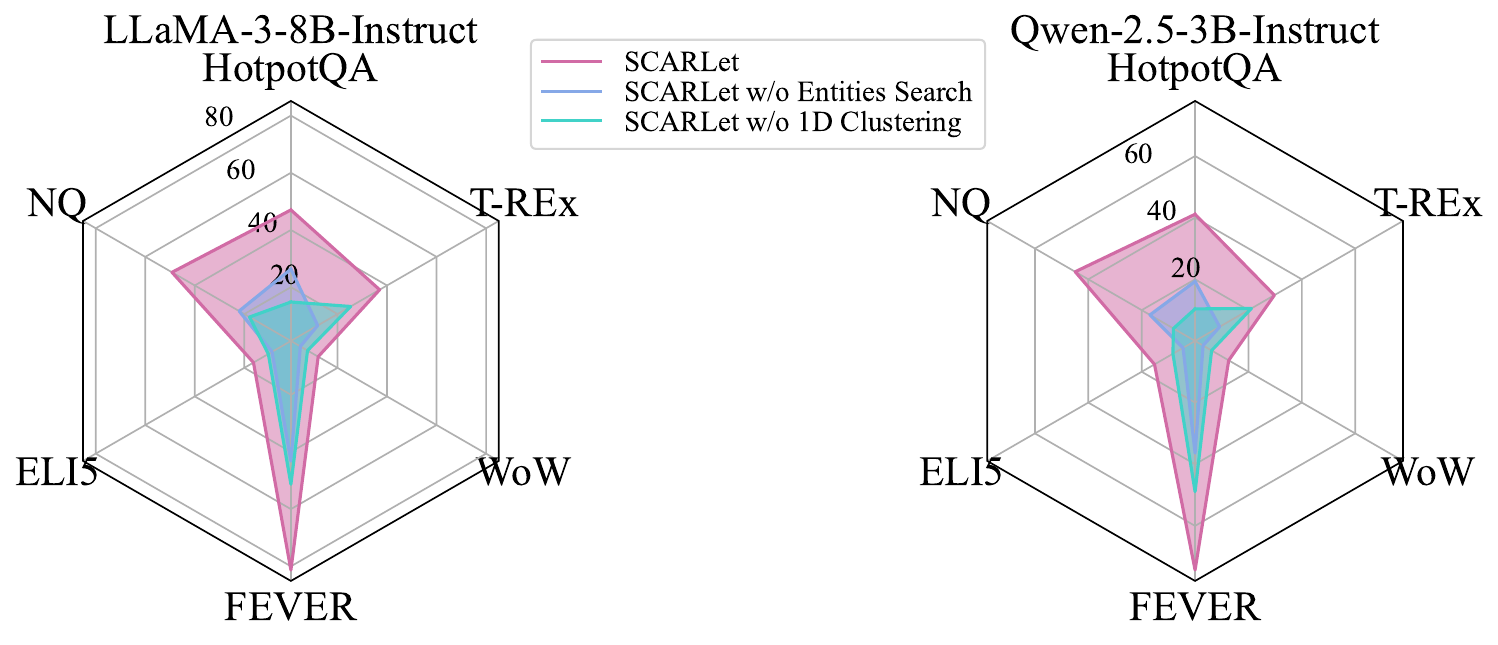}
  \caption {Ablation Study on six in-domain datasets, using BGE as retriever, with two generators. The values in the charts correspond to the metrics of each dataset.}
  \label{fig:fig-4}
\end{figure}

The comparison results, presented in Figure \ref{fig:fig-4}, show that removing either of the two components leads to a significant performance drop. Without adjacent entities retrieval, we believe that the original entity list may contain insufficient information, making it challenging to construct a shared context that effectively supports multi-task data synthesis. And the weaker entity associations can disrupt the connection between peer passages in the shared context, ultimately degrading the quality of the synthesized data. Furthermore, without one-dimensional clustering sampling, we suggest that it reduces the number of positive samples, which can be particularly detrimental to retrieval tasks requiring multiple reasoning hops.

\subsection{Aspects of Retrieval Utility}
\label{sec:5-3}

The previous experiment evaluates the overall performance improvement of RALMs brought by SCARLet. However, in essence, SCARLet is an optimization method of the retrieval stage. Moreover, despite discussing the utility as the valid gain for downstream generation in RALMs, neither existing work nor this study can explicitly define utility-based retrieval. To assess the effectiveness of SCARLet in improving retrieval performance, we select three retrieval benchmarks, each representing a distinct aspect of retrieval utility based on our understanding, as shown below:

\paragraph{GTI} This benchmark was introduced in Section \ref{sec:3-3}. Its goal is to evaluate whether retrievers can bypass pitfalls of semantic relevance and prioritize passages that are useful for answering questions.

\paragraph{\textsc{Bright}} This benchmark focuses on the reasoning implied in retrieval ~\citep{bright}, particularly for complex queries that require the retriever to engage in deep reasoning to identify useful passages, beyond simple semantic relevance. \citet{dai2024improvedensepassageretrieval} also argue that the entailment reasoning between passages and queries is essential for enhancing retrieval capabilities. We believe that recognizing retrieval utility requires reasoning, such as distinguishing task-specific features and determining the appropriate number of hops.

\paragraph{$\mathbb{X}^2$-Retrieval} This benchmark focuses on retrieval across multiple tasks and scenarios ~\citep{tart}, where understanding the intent behind user's queries becomes crucial. We suggest that this corresponds to identifying the target utility anticipated by the downstream tasks.

We choose Contriever and BGE as the retriever models, using LLaMA-3-8B-Instruct as the downstream generator to implement SCARLet training. We compare the performance of the trained retrievers with the initial retrievers on two benchmarks, as shown in Table \ref{tab:tb-3}, \ref{tab:tb-4} and \ref{tab:tb-5}, respectively. The results indicate that SCARLet improves performance on some datasets, but its effectiveness is generally limited for code-related tasks, such as LinkSo~\citep{linkso} and CodeSearchNet~\citep{codesearchnet}. The reasons could be: 1) the significant difference between the code domain and our selected in-domain datasets, which may hinder generalization; 2) the retriever models used are relatively lightweight, making it susceptible to catastrophic forgetting during training; 3) the optimization is related to downstream generators, but feedback related to the code domain cannot be obtained.

\begin{table}[t!]
\centering
\resizebox{\linewidth}{!}{
\renewcommand\arraystretch{1.1}
\begin{tabular}{lcccccc}
    \toprule
    \multirow{3}{*}{\textbf{Method}} & \multicolumn{2}{c}{\textbf{HotpotQA}} & \multicolumn{2}{c}{\textbf{NQ}} & \multicolumn{2}{c}{\textbf{MSMARCO-QA}} \\
    \cmidrule(lr){2-3} \cmidrule(lr){4-5} \cmidrule(lr){6-7}
    & \textbf{NDCG@1} & \textbf{NDCG@5} & \textbf{NDCG@1} & \textbf{NDCG@5} & \textbf{NDCG@1} & \textbf{NDCG@5} \\
    \midrule
    Contriever & 33.3 & 48.0 & 10.0 & 35.8 & 16.8 & 37.0 \\
    \rowcolor{mypurple}
    $\textrm{SCARLet}_{\textrm{Contriever}}$ & $\textrm{41.3}_{\textcolor{mygreen}{\left( +8.0 \right)}}$ & $\textrm{52.1}_{\textcolor{mygreen}{\left( +4.1 \right)}}$ & $\textrm{17.5}_{\textcolor{mygreen}{\left( +7.5 \right)}}$ & $\textrm{45.3}_{\textcolor{mygreen}{\left( +9.5 \right)}}$ & $\textrm{21.9}_{\textcolor{mygreen}{\left( +5.1 \right)}}$ & $\textrm{44.1}_{\textcolor{mygreen}{\left( +7.1 \right)}}$\\
    \midrule
    BGE & 70.3 & 70.1 & 30.3 & 60.2 & 47.8 & 71.9 \\
    \rowcolor{mypurple}
    $\textrm{SCARLet}_{\textrm{BGE}}$ & $\textrm{72.8}_{\textcolor{mygreen}{\left( +2.5 \right)}}$ & $\textrm{76.7}_{\textcolor{mygreen}{\left( +6.6 \right)}}$ & $\textrm{33.4}_{\textcolor{mygreen}{\left( +3.1 \right)}}$ & $\textrm{64.4}_{\textcolor{mygreen}{\left( +4.2 \right)}}$ & $\textrm{53.2}_{\textcolor{mygreen}{\left( +5.4 \right)}}$ & $\textrm{77.0}_{\textcolor{mygreen}{\left( +5.1 \right)}}$\\
    \bottomrule
\end{tabular}
}
\caption{Evaluation results on GTI, reporting nDCG for each datasets. Bracketed values indicate the changes in metrics compared to the initial model.}
\label{tab:tb-3}
\end{table}

\begin{table}[t!]
\centering
\resizebox{\linewidth}{!}{
\renewcommand\arraystretch{1.1}
\begin{tabular}{lccc}
    \toprule
    \textbf{Model} & \textbf{StackExchange} & \textbf{Coding} & \textbf{Theorem-based} \\
    \midrule
    Contriever & 10.5 & 19.6 & 6.9 \\
    \rowcolor{mypurple}
    $\textrm{SCARLet}_{\textrm{Contriever}}$ & $\textrm{13.3}_{\textcolor{mygreen}{\left( +2.8 \right)}}$  & $\textrm{19.2}_{\textcolor{myred}{\left( -0.4 \right)}}$ & $\textrm{8.7}_{\textcolor{mygreen}{\left( +1.8 \right)}}$ \\
    \midrule
    BGE & 14.9 & 16.0 & 8.1 \\
    \rowcolor{mypurple}
    $\textrm{SCARLet}_{\textrm{BGE}}$ & $\textrm{16.2}_{\textcolor{mygreen}{\left( +1.3 \right)}}$  & $\textrm{14.4}_{\textcolor{myred}{\left( -1.6 \right)}}$ & $\textrm{9.2}_{\textcolor{mygreen}{\left( +1.1 \right)}}$ \\
    \bottomrule
\end{tabular}
}
\caption{Evaluation results on \textsc{Bright}, reporting nDCG@10 for each datasets. Bracketed values indicate the changes in metrics compared to the initial model.}
\label{tab:tb-4}
\end{table}

\begin{table}[t!]
\centering
\resizebox{\linewidth}{!}{
\renewcommand\arraystretch{1.1}
\begin{tabular}{lccccc}
    \toprule
    \textbf{Model} & \textbf{AMB} & \textbf{WQA} & \textbf{GAT} & \textbf{LSO} & \textbf{CSP} \\
    \midrule
    Contriever & 96.8  & 80.9  & 73.2 & 28.0 & 36.7 \\
    \rowcolor{mypurple}
    $\textrm{SCARLet}_{\textrm{Contriever}}$ & $\textrm{97.5}_{\textcolor{mygreen}{\left( +0.7 \right)}}$ & $\textrm{85.8}_{\textcolor{mygreen}{\left( +5.1 \right)}}$ & $\textrm{71.6}_{\textcolor{myred}{\left( -1.6 \right)}}$ & $\textrm{20.9}_{\textcolor{myred}{\left( -7.1 \right)}}$ & $\textrm{24.8}_{\textcolor{myred}{\left( -11.9 \right)}}$ \\
    \midrule
    BGE & 97.3 & 84.0 & 77.4 & 30.1 & 38.2  \\
    \rowcolor{mypurple}
    $\textrm{SCARLet}_{\textrm{BGE}}$ & $\textrm{98.3}_{\textcolor{mygreen}{\left( +1.0 \right)}}$ & $\textrm{86.1}_{\textcolor{mygreen}{\left( +2.1 \right)}}$ & $\textrm{77.8}_{\textcolor{mygreen}{\left( +0.4 \right)}}$ & $\textrm{27.5}_{\textcolor{myred}{\left( -2.6 \right)}}$ & $\textrm{34.9}_{\textcolor{myred}{\left( -3.3 \right)}}$ \\
    \bottomrule
\end{tabular}
}
\caption{Evaluation results on $\mathbb{X}^2$-Retrieval, averaged nDCG@10 for each datasets. Bracketed values indicate the changes in metrics compared to the initial model.}
\label{tab:tb-5}
\end{table}

\subsection{Case Study}
\label{sec:5-4}

\begin{figure}[htbp]
  \includegraphics[width=\linewidth,scale=1.00]{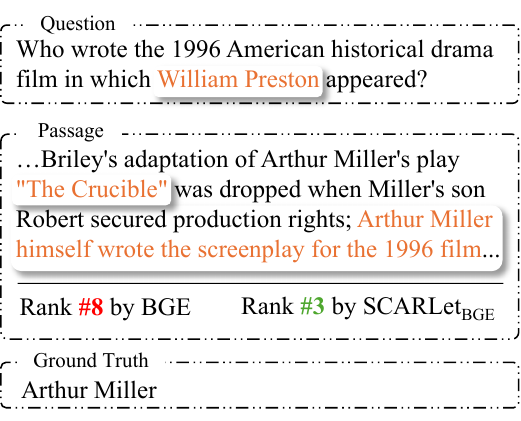}
  \caption {Case Study on HotpotQA. The passage is ranked variously by different retrievers. Orange text indicates necessary reasoning information. }
  \label{fig:fig-5}
\end{figure}

Multi-hop QA is a task that requires multiple pieces of information and multi-step reasoning to derive the solution~\citep{mavi2024multihopquestionanswering}. Given the characteristics of the task, we believe that retrieval utility should point to passages that may contain information necessary for the reasoning chain. We select a representative example from the test split of the HotpotQA dataset, as shown in Figure \ref{fig:fig-5}. To answer the question, the reasoning chain is: knowing information about William Preston, identifying the 1996 American historical drama he appeared in, finding information about that drama, and determining its writer. Directly relevant information about William Preston is relatively easy to define. However, the shown passage which corresponds to the final reasoning step, has a poor match with the question in terms of semantic relevance. And BGE ranks it 8th. After training by SCARLet, the passage achieves a higher ranking of 3rd. For more case studies, please refer to Appendix \ref{sec:appendix_e}.

\section{Conclusion}

This study focuses on utility-based retrieval, a paradigm that moves beyond semantic relevance to prioritize downstream task performance in RALMs. We highlight two key challenges faced by existing research. To solve the limitations, we propose SCARLet, a novel framework to enhance utility-based retrieval. To mitigate semantic interference on utility features during training, SCARLet incorporates a shared context synthesis method, which narrows the semantic gap between different tasks. To address the issue of inaccurate passage-level utility estimation, SCARLet employs a perturbation-based attribution method to capture the synergy between passages. Lastly, SCARLet utilizes a one-dimensional clustering method to sample positive and negative passages for retriever optimization. Through experiments, we demonstrate that SCARLet can effectively enhance the overall performance of RALMs, and brings improvements in complex retrieval benchmarks. We hope this study can inspire further research on utility-based retrieval.

\section*{Limitations}

This study only covers several classic downstream datasets. We believe that incorporating a task augmentation stage could further enhance generalization, which we leave for future work. Moreover, there is a noticeable decline in retrieval performance in the code domain during generalization tests. Therefore, future work should also focus on improving the integration of different corpus structures. In addition, due to environmental constraints, this study does not evaluate larger-scale retrievers and generators. Furthermore, we also try GPT-4o-mini as the synthesizer, which performed poorly. Thus our framework should be equipped with models with stronger reasoning capabilities.

\section*{Ethics Statement}

All datasets and corpora involved in this study are publicly available, and we ensure that all used data comply with the usage and privacy policies established by the original authors. The synthetic data is exclusively used for training the retriever model. Moreover, given the security assurance of the synthesizer model, the probability of generating harmful passages and data is extremely minimal.

\section*{Acknowledgments}

This work was supported by the National Key R\&D Program of China (2023YFC3303800).

\bibliography{custom}

\begin{thebibliography}{72}
\providecommand{\natexlab}[1]{#1}

\bibitem[{AI@Meta(2024)}]{llama3modelcard}
AI@Meta. 2024.
\newblock \href {https://github.com/meta-llama/llama3/blob/main/MODEL_CARD.md} {Llama 3 model card}.

\bibitem[{Asai et~al.(2023)Asai, Schick, Lewis, Chen, Izacard, Riedel, Hajishirzi, and Yih}]{tart}
Akari Asai, Timo Schick, Patrick Lewis, Xilun Chen, Gautier Izacard, Sebastian Riedel, Hannaneh Hajishirzi, and Wen-tau Yih. 2023.
\newblock \href {https://doi.org/10.18653/v1/2023.findings-acl.225} {Task-aware retrieval with instructions}.
\newblock In \emph{Findings of the Association for Computational Linguistics: ACL 2023}, pages 3650--3675, Toronto, Canada. Association for Computational Linguistics.

\bibitem[{Bajaj et~al.(2018)Bajaj, Campos, Craswell, Deng, Gao, Liu, Majumder, McNamara, Mitra, Nguyen, Rosenberg, Song, Stoica, Tiwary, and Wang}]{msmarcoqa}
Payal Bajaj, Daniel Campos, Nick Craswell, Li~Deng, Jianfeng Gao, Xiaodong Liu, Rangan Majumder, Andrew McNamara, Bhaskar Mitra, Tri Nguyen, Mir Rosenberg, Xia Song, Alina Stoica, Saurabh Tiwary, and Tong Wang. 2018.
\newblock \href {https://arxiv.org/abs/1611.09268} {Ms marco: A human generated machine reading comprehension dataset}.
\newblock \emph{Preprint}, arXiv:1611.09268.

\bibitem[{Bi et~al.(2024)Bi, Liu, Wang, Mei, Gao, Xu, and Cheng}]{bi2024adaptivetokenbiaserknowledge}
Baolong Bi, Shenghua Liu, Yiwei Wang, Lingrui Mei, Hongcheng Gao, Yilong Xu, and Xueqi Cheng. 2024.
\newblock \href {https://arxiv.org/abs/2406.12468} {Adaptive token biaser: Knowledge editing via biasing key entities}.
\newblock \emph{Preprint}, arXiv:2406.12468.

\bibitem[{Bi et~al.(2025)Bi, Liu, Wang, Xu, Fang, Mei, and Cheng}]{bi2025parameters}
Baolong Bi, Shenghua Liu, Yiwei Wang, Yilong Xu, Junfeng Fang, Lingrui Mei, and Xueqi Cheng. 2025.
\newblock Parameters vs. context: Fine-grained control of knowledge reliance in language models.
\newblock \emph{arXiv preprint arXiv:2503.15888}.

\bibitem[{Brown et~al.(2020)Brown, Mann, Ryder, Subbiah, Kaplan, Dhariwal, Neelakantan, Shyam, Sastry, Askell et~al.}]{llm}
Tom Brown, Benjamin Mann, Nick Ryder, Melanie Subbiah, Jared~D Kaplan, Prafulla Dhariwal, Arvind Neelakantan, Pranav Shyam, Girish Sastry, Amanda Askell, et~al. 2020.
\newblock \href {https://proceedings.neurips.cc/paper_files/paper/2020/file/1457c0d6bfcb4967418bfb8ac142f64a-Paper.pdf} {Language models are few-shot learners}.
\newblock In \emph{Advances in Neural Information Processing Systems}, volume~33, pages 1877--1901. Curran Associates, Inc.

\bibitem[{Chen and Shu(2024)}]{chen2024llmgeneratedmisinformationdetected}
Canyu Chen and Kai Shu. 2024.
\newblock \href {https://arxiv.org/abs/2309.13788} {Can llm-generated misinformation be detected?}
\newblock \emph{Preprint}, arXiv:2309.13788.

\bibitem[{Cheng et~al.(2023)Cheng, Luo, Chen, Liu, Zhao, and Yan}]{lift_yourself_up}
Xin Cheng, Di~Luo, Xiuying Chen, Lemao Liu, Dongyan Zhao, and Rui Yan. 2023.
\newblock \href {https://proceedings.neurips.cc/paper_files/paper/2023/file/887262aeb3eafb01ef0fd0e3a87a8831-Paper-Conference.pdf} {Lift yourself up: Retrieval-augmented text generation with self-memory}.
\newblock In \emph{Advances in Neural Information Processing Systems}, volume~36, pages 43780--43799. Curran Associates, Inc.

\bibitem[{Dai et~al.(2024)Dai, Liu, and Xiong}]{dai2024improvedensepassageretrieval}
Lu~Dai, Hao Liu, and Hui Xiong. 2024.
\newblock \href {https://arxiv.org/abs/2410.15801} {Improve dense passage retrieval with entailment tuning}.
\newblock \emph{Preprint}, arXiv:2410.15801.

\bibitem[{Denil et~al.(2015)Denil, Demiraj, and de~Freitas}]{gtimesi}
Misha Denil, Alban Demiraj, and Nando de~Freitas. 2015.
\newblock \href {https://arxiv.org/abs/1412.6815} {Extraction of salient sentences from labelled documents}.
\newblock \emph{Preprint}, arXiv:1412.6815.

\bibitem[{Diggelmann et~al.(2021)Diggelmann, Boyd-Graber, Bulian, Ciaramita, and Leippold}]{diggelmann2021climatefeverdatasetverificationrealworld}
Thomas Diggelmann, Jordan Boyd-Graber, Jannis Bulian, Massimiliano Ciaramita, and Markus Leippold. 2021.
\newblock \href {https://arxiv.org/abs/2012.00614} {Climate-fever: A dataset for verification of real-world climate claims}.
\newblock \emph{Preprint}, arXiv:2012.00614.

\bibitem[{Dinan et~al.(2019)Dinan, Roller, Shuster, Fan, Auli, and Weston}]{dinan2019wizard}
Emily Dinan, Stephen Roller, Kurt Shuster, Angela Fan, Michael Auli, and Jason Weston. 2019.
\newblock {W}izard of {W}ikipedia: Knowledge-powered conversational agents.
\newblock In \emph{Proceedings of the International Conference on Learning Representations (ICLR)}.

\bibitem[{Elsahar et~al.(2018)Elsahar, Vougiouklis, Remaci, Gravier, Hare, Laforest, and Simperl}]{elsahar-etal-2018-trex}
Hady Elsahar, Pavlos Vougiouklis, Arslen Remaci, Christophe Gravier, Jonathon Hare, Frederique Laforest, and Elena Simperl. 2018.
\newblock \href {https://aclanthology.org/L18-1544/} {{T}-{RE}x: A large scale alignment of natural language with knowledge base triples}.
\newblock In \emph{Proceedings of the Eleventh International Conference on Language Resources and Evaluation ({LREC} 2018)}, Miyazaki, Japan. European Language Resources Association (ELRA).

\bibitem[{Fan et~al.(2019)Fan, Jernite, Perez, Grangier, Weston, and Auli}]{fan-etal-2019-eli5}
Angela Fan, Yacine Jernite, Ethan Perez, David Grangier, Jason Weston, and Michael Auli. 2019.
\newblock \href {https://doi.org/10.18653/v1/P19-1346} {{ELI}5: Long form question answering}.
\newblock In \emph{Proceedings of the 57th Annual Meeting of the Association for Computational Linguistics}, pages 3558--3567, Florence, Italy. Association for Computational Linguistics.

\bibitem[{Fang et~al.(2024)Fang, Bai, Ni, Yang, Chen, and Xu}]{raat}
Feiteng Fang, Yuelin Bai, Shiwen Ni, Min Yang, Xiaojun Chen, and Ruifeng Xu. 2024.
\newblock \href {https://arxiv.org/abs/2405.20978} {Enhancing noise robustness of retrieval-augmented language models with adaptive adversarial training}.
\newblock \emph{Preprint}, arXiv:2405.20978.

\bibitem[{Gao et~al.(2024)Gao, Xiong, Gao, Jia, Pan, Bi, Dai, Sun, Wang, and Wang}]{ragsurvey}
Yunfan Gao, Yun Xiong, Xinyu Gao, Kangxiang Jia, Jinliu Pan, Yuxi Bi, Yi~Dai, Jiawei Sun, Meng Wang, and Haofen Wang. 2024.
\newblock \href {https://arxiv.org/abs/2312.10997} {Retrieval-augmented generation for large language models: A survey}.
\newblock \emph{Preprint}, arXiv:2312.10997.

\bibitem[{Glass et~al.(2022)Glass, Rossiello, Chowdhury, Naik, Cai, and Gliozzo}]{glass-etal-2022-re2g}
Michael Glass, Gaetano Rossiello, Md~Faisal~Mahbub Chowdhury, Ankita Naik, Pengshan Cai, and Alfio Gliozzo. 2022.
\newblock \href {https://doi.org/10.18653/v1/2022.naacl-main.194} {{R}e2{G}: Retrieve, rerank, generate}.
\newblock In \emph{Proceedings of the 2022 Conference of the North American Chapter of the Association for Computational Linguistics: Human Language Technologies}, pages 2701--2715, Seattle, United States. Association for Computational Linguistics.

\bibitem[{Guo et~al.(2019)Guo, Fan, Ji, and Cheng}]{matchzoo}
Jiafeng Guo, Yixing Fan, Xiang Ji, and Xueqi Cheng. 2019.
\newblock \href {https://doi.org/10.1145/3331184.3331403} {Matchzoo: A learning, practicing, and developing system for neural text matching}.
\newblock In \emph{Proceedings of the 42nd International ACM SIGIR Conference on Research and Development in Information Retrieval}, SIGIR'19, page 1297–1300, New York, NY, USA. Association for Computing Machinery.

\bibitem[{Han and Tsvetkov(2022)}]{pretrainattribution1}
Xiaochuang Han and Yulia Tsvetkov. 2022.
\newblock \href {https://arxiv.org/abs/2205.12600} {Orca: Interpreting prompted language models via locating supporting data evidence in the ocean of pretraining data}.
\newblock \emph{Preprint}, arXiv:2205.12600.

\bibitem[{Hilt and Seegrist(1977)}]{Hilt1977RidgeAC}
Donald~E. Hilt and Donald~W. Seegrist. 1977.
\newblock \href {https://api.semanticscholar.org/CorpusID:106850190} {Ridge: a computer program for calculating ridge regression estimates}.

\bibitem[{Huang et~al.(2024)Huang, Yu, Ma, Zhong, Feng, Wang, Chen, Peng, Feng, Qin, and Liu}]{hallucination1}
Lei Huang, Weijiang Yu, Weitao Ma, Weihong Zhong, Zhangyin Feng, Haotian Wang, Qianglong Chen, Weihua Peng, Xiaocheng Feng, Bing Qin, and Ting Liu. 2024.
\newblock \href {https://doi.org/10.1145/3703155} {A survey on hallucination in large language models: Principles, taxonomy, challenges, and open questions}.
\newblock \emph{ACM Transactions on Information Systems}.

\bibitem[{Husain et~al.(2020)Husain, Wu, Gazit, Allamanis, and Brockschmidt}]{codesearchnet}
Hamel Husain, Ho-Hsiang Wu, Tiferet Gazit, Miltiadis Allamanis, and Marc Brockschmidt. 2020.
\newblock \href {https://arxiv.org/abs/1909.09436} {Codesearchnet challenge: Evaluating the state of semantic code search}.
\newblock \emph{Preprint}, arXiv:1909.09436.

\bibitem[{Izacard et~al.(2022)Izacard, Caron, Hosseini, Riedel, Bojanowski, Joulin, and Grave}]{contriever}
Gautier Izacard, Mathilde Caron, Lucas Hosseini, Sebastian Riedel, Piotr Bojanowski, Armand Joulin, and Edouard Grave. 2022.
\newblock \href {https://arxiv.org/abs/2112.09118} {Unsupervised dense information retrieval with contrastive learning}.
\newblock \emph{Preprint}, arXiv:2112.09118.

\bibitem[{Karpukhin et~al.(2020)Karpukhin, Oguz, Min, Lewis, Wu, Edunov, Chen, and Yih}]{dpr}
Vladimir Karpukhin, Barlas Oguz, Sewon Min, Patrick Lewis, Ledell Wu, Sergey Edunov, Danqi Chen, and Wen-tau Yih. 2020.
\newblock \href {https://doi.org/10.18653/v1/2020.emnlp-main.550} {Dense passage retrieval for open-domain question answering}.
\newblock In \emph{Proceedings of the 2020 Conference on Empirical Methods in Natural Language Processing (EMNLP)}, pages 6769--6781, Online. Association for Computational Linguistics.

\bibitem[{Kim and Baek(2025)}]{kim2025syntrievertrainretrieversynthetic}
Minsang Kim and Seungjun Baek. 2025.
\newblock \href {https://arxiv.org/abs/2502.03824} {Syntriever: How to train your retriever with synthetic data from llms}.
\newblock \emph{Preprint}, arXiv:2502.03824.

\bibitem[{Kwiatkowski et~al.(2019)Kwiatkowski, Palomaki, Redfield, Collins, Parikh, Alberti, Epstein, Polosukhin, Devlin, Lee, Toutanova, Jones, Kelcey, Chang, Dai, Uszkoreit, Le, and Petrov}]{kwiatkowski-etal-2019-nq}
Tom Kwiatkowski, Jennimaria Palomaki, Olivia Redfield, Michael Collins, Ankur Parikh, Chris Alberti, Danielle Epstein, Illia Polosukhin, Jacob Devlin, Kenton Lee, Kristina Toutanova, Llion Jones, Matthew Kelcey, Ming-Wei Chang, Andrew~M. Dai, Jakob Uszkoreit, Quoc Le, and Slav Petrov. 2019.
\newblock \href {https://doi.org/10.1162/tacl_a_00276} {Natural questions: A benchmark for question answering research}.
\newblock \emph{Transactions of the Association for Computational Linguistics}, 7:452--466.

\bibitem[{Levy et~al.(2017)Levy, Seo, Choi, and Zettlemoyer}]{levy-etal-2017-zero}
Omer Levy, Minjoon Seo, Eunsol Choi, and Luke Zettlemoyer. 2017.
\newblock \href {https://doi.org/10.18653/v1/K17-1034} {Zero-shot relation extraction via reading comprehension}.
\newblock In \emph{Proceedings of the 21st Conference on Computational Natural Language Learning ({C}o{NLL} 2017)}, pages 333--342, Vancouver, Canada. Association for Computational Linguistics.

\bibitem[{Lewis et~al.(2020)Lewis, Perez, Piktus, Petroni, Karpukhin, Goyal, K\"{u}ttler, Lewis, Yih, Rockt\"{a}schel, Riedel, and Kiela}]{rag_lewis_2020}
Patrick Lewis, Ethan Perez, Aleksandra Piktus, Fabio Petroni, Vladimir Karpukhin, Naman Goyal, Heinrich K\"{u}ttler, Mike Lewis, Wen-tau Yih, Tim Rockt\"{a}schel, Sebastian Riedel, and Douwe Kiela. 2020.
\newblock \href {https://proceedings.neurips.cc/paper_files/paper/2020/file/6b493230205f780e1bc26945df7481e5-Paper.pdf} {Retrieval-augmented generation for knowledge-intensive nlp tasks}.
\newblock In \emph{Advances in Neural Information Processing Systems}, volume~33, pages 9459--9474. Curran Associates, Inc.

\bibitem[{Li et~al.(2023)Li, Sun, Hu, Liu, Chen, Hu, Wu, and Zhang}]{li2023attributionsurvey}
Dongfang Li, Zetian Sun, Xinshuo Hu, Zhenyu Liu, Ziyang Chen, Baotian Hu, Aiguo Wu, and Min Zhang. 2023.
\newblock \href {https://arxiv.org/abs/2311.03731} {A survey of large language models attribution}.
\newblock \emph{Preprint}, arXiv:2311.03731.

\bibitem[{Li et~al.(2024)Li, Cao, Pan, Ma, and Sun}]{ragattribution2}
Xinze Li, Yixin Cao, Liangming Pan, Yubo Ma, and Aixin Sun. 2024.
\newblock \href {https://arxiv.org/abs/2310.05634} {Towards verifiable generation: A benchmark for knowledge-aware language model attribution}.
\newblock \emph{Preprint}, arXiv:2310.05634.

\bibitem[{Lin et~al.(2024)Lin, Chen, Chen, Shi, Lomeli, James, Rodriguez, Kahn, Szilvasy, Lewis, Zettlemoyer, and Yih}]{lin2024raditretrievalaugmenteddualinstruction}
Xi~Victoria Lin, Xilun Chen, Mingda Chen, Weijia Shi, Maria Lomeli, Rich James, Pedro Rodriguez, Jacob Kahn, Gergely Szilvasy, Mike Lewis, Luke Zettlemoyer, and Scott Yih. 2024.
\newblock \href {https://arxiv.org/abs/2310.01352} {Ra-dit: Retrieval-augmented dual instruction tuning}.
\newblock \emph{Preprint}, arXiv:2310.01352.

\bibitem[{Liu et~al.(2018)Liu, Wang, Leng, and Zhai}]{linkso}
Xueqing Liu, Chi Wang, Yue Leng, and ChengXiang Zhai. 2018.
\newblock \href {https://doi.org/10.1145/3283812.3283815} {Linkso: a dataset for learning to retrieve similar question answer pairs on software development forums}.
\newblock In \emph{Proceedings of the 4th ACM SIGSOFT International Workshop on NLP for Software Engineering}, NL4SE 2018, page 2–5, New York, NY, USA. Association for Computing Machinery.

\bibitem[{Liu et~al.(2024)Liu, Hu, Zhang, Chen, Wu, and Wu}]{liu2024finegrainedguidanceretrieversleveraging}
Yuhang Liu, Xueyu Hu, Shengyu Zhang, Jingyuan Chen, Fan Wu, and Fei Wu. 2024.
\newblock \href {https://arxiv.org/abs/2411.03957} {Fine-grained guidance for retrievers: Leveraging llms' feedback in retrieval-augmented generation}.
\newblock \emph{Preprint}, arXiv:2411.03957.

\bibitem[{Long et~al.(2024)Long, Wang, Xiao, Zhao, Ding, Chen, and Wang}]{long2024llmsdrivensyntheticdatageneration}
Lin Long, Rui Wang, Ruixuan Xiao, Junbo Zhao, Xiao Ding, Gang Chen, and Haobo Wang. 2024.
\newblock \href {https://arxiv.org/abs/2406.15126} {On llms-driven synthetic data generation, curation, and evaluation: A survey}.
\newblock \emph{Preprint}, arXiv:2406.15126.

\bibitem[{Lopardo et~al.(2024)Lopardo, Precioso, and Garreau}]{attentionmeetsposthocinterpretability}
Gianluigi Lopardo, Frederic Precioso, and Damien Garreau. 2024.
\newblock \href {https://proceedings.mlr.press/v235/lopardo24a.html} {Attention meets post-hoc interpretability: A mathematical perspective}.
\newblock In \emph{Proceedings of the 41st International Conference on Machine Learning}, volume 235 of \emph{Proceedings of Machine Learning Research}, pages 32781--32800. PMLR.

\bibitem[{Madsen et~al.(2024)Madsen, Chandar, and Reddy}]{madsen-etal-2024-self}
Andreas Madsen, Sarath Chandar, and Siva Reddy. 2024.
\newblock \href {https://doi.org/10.18653/v1/2024.findings-acl.19} {Are self-explanations from large language models faithful?}
\newblock In \emph{Findings of the Association for Computational Linguistics: ACL 2024}, pages 295--337, Bangkok, Thailand. Association for Computational Linguistics.

\bibitem[{Maia et~al.(2018)Maia, Handschuh, Freitas, Davis, McDermott, Zarrouk, and Balahur}]{fiqa}
Macedo Maia, Siegfried Handschuh, Andr\'{e} Freitas, Brian Davis, Ross McDermott, Manel Zarrouk, and Alexandra Balahur. 2018.
\newblock \href {https://doi.org/10.1145/3184558.3192301} {Www'18 open challenge: Financial opinion mining and question answering}.
\newblock In \emph{Companion Proceedings of the The Web Conference 2018}, WWW '18, page 1941–1942, Republic and Canton of Geneva, CHE. International World Wide Web Conferences Steering Committee.

\bibitem[{Mardaoui and Garreau(2021)}]{ananalysisoflimefortextdata}
Dina Mardaoui and Damien Garreau. 2021.
\newblock \href {https://proceedings.mlr.press/v130/mardaoui21a.html} {An analysis of lime for text data}.
\newblock In \emph{Proceedings of The 24th International Conference on Artificial Intelligence and Statistics}, volume 130 of \emph{Proceedings of Machine Learning Research}, pages 3493--3501. PMLR.

\bibitem[{Mavi et~al.(2024)Mavi, Jangra, and Jatowt}]{mavi2024multihopquestionanswering}
Vaibhav Mavi, Anubhav Jangra, and Adam Jatowt. 2024.
\newblock \href {https://arxiv.org/abs/2204.09140} {Multi-hop question answering}.
\newblock \emph{Preprint}, arXiv:2204.09140.

\bibitem[{Mylonas et~al.(2022)Mylonas, Mollas, and Tsoumakas}]{mylonas2022attentionmatrixdecisionfaithfulnessbased}
Nikolaos Mylonas, Ioannis Mollas, and Grigorios Tsoumakas. 2022.
\newblock \href {https://arxiv.org/abs/2209.10876} {An attention matrix for every decision: Faithfulness-based arbitration among multiple attention-based interpretations of transformers in text classification}.
\newblock \emph{Preprint}, arXiv:2209.10876.

\bibitem[{Naveed et~al.(2024)Naveed, Khan, Qiu, Saqib, Anwar, Usman, Akhtar, Barnes, and Mian}]{llmsurvey2}
Humza Naveed, Asad~Ullah Khan, Shi Qiu, Muhammad Saqib, Saeed Anwar, Muhammad Usman, Naveed Akhtar, Nick Barnes, and Ajmal Mian. 2024.
\newblock \href {https://arxiv.org/abs/2307.06435} {A comprehensive overview of large language models}.
\newblock \emph{Preprint}, arXiv:2307.06435.

\bibitem[{Nielsen et~al.(2022)Nielsen, Dera, Rasool, Ramachandran, and Bouaynaya}]{gradientbasedattributionmethods}
Ian~E. Nielsen, Dimah Dera, Ghulam Rasool, Ravi~P. Ramachandran, and Nidhal~Carla Bouaynaya. 2022.
\newblock \href {https://doi.org/10.1109/MSP.2022.3142719} {Robust explainability: A tutorial on gradient-based attribution methods for deep neural networks}.
\newblock \emph{IEEE Signal Processing Magazine}, 39(4):73--84.

\bibitem[{OpenAI(2024)}]{gpt4o}
OpenAI. 2024.
\newblock \href {https://openai.com/index/gpt-4o-system-card/} {Gpt-4o system card}.

\bibitem[{Petroni et~al.(2021)Petroni, Piktus, Fan, Lewis, Yazdani, De~Cao, Thorne, Jernite, Karpukhin, Maillard, Plachouras, Rockt{\"a}schel, and Riedel}]{petroni-etal-2021-kilt}
Fabio Petroni, Aleksandra Piktus, Angela Fan, Patrick Lewis, Majid Yazdani, Nicola De~Cao, James Thorne, Yacine Jernite, Vladimir Karpukhin, Jean Maillard, Vassilis Plachouras, Tim Rockt{\"a}schel, and Sebastian Riedel. 2021.
\newblock \href {https://doi.org/10.18653/v1/2021.naacl-main.200} {{KILT}: a benchmark for knowledge intensive language tasks}.
\newblock In \emph{Proceedings of the 2021 Conference of the North American Chapter of the Association for Computational Linguistics: Human Language Technologies}, pages 2523--2544, Online. Association for Computational Linguistics.

\bibitem[{Ribeiro et~al.(2016)Ribeiro, Singh, and Guestrin}]{lime}
Marco~Tulio Ribeiro, Sameer Singh, and Carlos Guestrin. 2016.
\newblock \href {https://arxiv.org/abs/1606.05386} {Model-agnostic interpretability of machine learning}.
\newblock \emph{Preprint}, arXiv:1606.05386.

\bibitem[{Robertson and Zaragoza(2009)}]{bm25}
Stephen Robertson and Hugo Zaragoza. 2009.
\newblock \href {https://doi.org/10.1561/1500000019} {The probabilistic relevance framework: Bm25 and beyond}.
\newblock \emph{Foundations and Trends® in Information Retrieval}, 3(4):333--389.

\bibitem[{Salemi and Zamani(2024)}]{towardsasearchengineformachines}
Alireza Salemi and Hamed Zamani. 2024.
\newblock \href {https://doi.org/10.1145/3626772.3657733} {Towards a search engine for machines: Unified ranking for multiple retrieval-augmented large language models}.
\newblock In \emph{Proceedings of the 47th International ACM SIGIR Conference on Research and Development in Information Retrieval}, SIGIR '24, page 741–751, New York, NY, USA. Association for Computing Machinery.

\bibitem[{Shao et~al.(2023)Shao, Gong, Shen, Huang, Duan, and Chen}]{shao2023enhancingretrievalaugmentedlargelanguage}
Zhihong Shao, Yeyun Gong, Yelong Shen, Minlie Huang, Nan Duan, and Weizhu Chen. 2023.
\newblock \href {https://arxiv.org/abs/2305.15294} {Enhancing retrieval-augmented large language models with iterative retrieval-generation synergy}.
\newblock \emph{Preprint}, arXiv:2305.15294.

\bibitem[{Shi et~al.(2023)Shi, Min, Yasunaga, Seo, James, Lewis, Zettlemoyer, and tau Yih}]{replug}
Weijia Shi, Sewon Min, Michihiro Yasunaga, Minjoon Seo, Rich James, Mike Lewis, Luke Zettlemoyer, and Wen tau Yih. 2023.
\newblock \href {https://arxiv.org/abs/2301.12652} {Replug: Retrieval-augmented black-box language models}.
\newblock \emph{Preprint}, arXiv:2301.12652.

\bibitem[{Shuster et~al.(2021)Shuster, Poff, Chen, Kiela, and Weston}]{ragattribution1}
Kurt Shuster, Spencer Poff, Moya Chen, Douwe Kiela, and Jason Weston. 2021.
\newblock \href {https://doi.org/10.18653/v1/2021.findings-emnlp.320} {Retrieval augmentation reduces hallucination in conversation}.
\newblock In \emph{Findings of the Association for Computational Linguistics: EMNLP 2021}, pages 3784--3803, Punta Cana, Dominican Republic. Association for Computational Linguistics.

\bibitem[{Sohn et~al.(2024)Sohn, Park, Yoon, Park, Hwang, Sung, Kim, and Kang}]{sohn2024rationaleguidedretrievalaugmentedgeneration}
Jiwoong Sohn, Yein Park, Chanwoong Yoon, Sihyeon Park, Hyeon Hwang, Mujeen Sung, Hyunjae Kim, and Jaewoo Kang. 2024.
\newblock \href {https://arxiv.org/abs/2411.00300} {Rationale-guided retrieval augmented generation for medical question answering}.
\newblock \emph{Preprint}, arXiv:2411.00300.

\bibitem[{Su et~al.(2024)Su, Yen, Xia, Shi, Muennighoff, yu~Wang, Liu, Shi, Siegel, Tang, Sun, Yoon, Arik, Chen, and Yu}]{bright}
Hongjin Su, Howard Yen, Mengzhou Xia, Weijia Shi, Niklas Muennighoff, Han yu~Wang, Haisu Liu, Quan Shi, Zachary~S. Siegel, Michael Tang, Ruoxi Sun, Jinsung Yoon, Sercan~O. Arik, Danqi Chen, and Tao Yu. 2024.
\newblock \href {https://arxiv.org/abs/2407.12883} {Bright: A realistic and challenging benchmark for reasoning-intensive retrieval}.
\newblock \emph{Preprint}, arXiv:2407.12883.

\bibitem[{Team(2024)}]{qwen2.5}
Qwen Team. 2024.
\newblock \href {https://qwenlm.github.io/blog/qwen2.5/} {Qwen2.5: A party of foundation models}.

\bibitem[{Thakur et~al.(2021)Thakur, Reimers, Rücklé, Srivastava, and Gurevych}]{thakur2021beirheterogenousbenchmarkzeroshot}
Nandan Thakur, Nils Reimers, Andreas Rücklé, Abhishek Srivastava, and Iryna Gurevych. 2021.
\newblock \href {https://arxiv.org/abs/2104.08663} {Beir: A heterogenous benchmark for zero-shot evaluation of information retrieval models}.
\newblock \emph{Preprint}, arXiv:2104.08663.

\bibitem[{Thorne et~al.(2018)Thorne, Vlachos, Christodoulopoulos, and Mittal}]{Thorne18Fever}
James Thorne, Andreas Vlachos, Christos Christodoulopoulos, and Arpit Mittal. 2018.
\newblock {FEVER}: a large-scale dataset for fact extraction and {VERification}.
\newblock In \emph{NAACL-HLT}.

\bibitem[{Wadden et~al.(2020)Wadden, Lin, Lo, Wang, van Zuylen, Cohan, and Hajishirzi}]{wadden-etal-2020-scifact}
David Wadden, Shanchuan Lin, Kyle Lo, Lucy~Lu Wang, Madeleine van Zuylen, Arman Cohan, and Hannaneh Hajishirzi. 2020.
\newblock \href {https://doi.org/10.18653/v1/2020.emnlp-main.609} {Fact or fiction: Verifying scientific claims}.
\newblock In \emph{Proceedings of the 2020 Conference on Empirical Methods in Natural Language Processing (EMNLP)}, pages 7534--7550, Online. Association for Computational Linguistics.

\bibitem[{Wang et~al.(2024)Wang, Zhang, Guo, and Shen}]{wang2024gradientbasedfeatureattribution}
Yongjie Wang, Tong Zhang, Xu~Guo, and Zhiqi Shen. 2024.
\newblock \href {https://arxiv.org/abs/2403.10415} {Gradient based feature attribution in explainable ai: A technical review}.
\newblock \emph{Preprint}, arXiv:2403.10415.

\bibitem[{Wang et~al.(2019)Wang, Ng, Ma, Nallapati, and Xiang}]{wang-etal-2019-multi}
Zhiguo Wang, Patrick Ng, Xiaofei Ma, Ramesh Nallapati, and Bing Xiang. 2019.
\newblock \href {https://doi.org/10.18653/v1/D19-1599} {Multi-passage {BERT}: A globally normalized {BERT} model for open-domain question answering}.
\newblock In \emph{Proceedings of the 2019 Conference on Empirical Methods in Natural Language Processing and the 9th International Joint Conference on Natural Language Processing (EMNLP-IJCNLP)}, pages 5878--5882, Hong Kong, China. Association for Computational Linguistics.

\bibitem[{Wei et~al.(2022)Wei, Tay, Bommasani, Raffel, Zoph, Borgeaud, Yogatama, Bosma, Zhou, Metzler, Chi, Hashimoto, Vinyals, Liang, Dean, and Fedus}]{wei2022emergentabilitieslargelanguage}
Jason Wei, Yi~Tay, Rishi Bommasani, Colin Raffel, Barret Zoph, Sebastian Borgeaud, Dani Yogatama, Maarten Bosma, Denny Zhou, Donald Metzler, Ed~H. Chi, Tatsunori Hashimoto, Oriol Vinyals, Percy Liang, Jeff Dean, and William Fedus. 2022.
\newblock \href {https://arxiv.org/abs/2206.07682} {Emergent abilities of large language models}.
\newblock \emph{Preprint}, arXiv:2206.07682.

\bibitem[{Wei et~al.(2024)Wei, Chen, and Meng}]{wei2024instructrag}
Zhepei Wei, Wei-Lin Chen, and Yu~Meng. 2024.
\newblock \href {https://openreview.net/forum?id=EY45IFgVC5} {Instruct{RAG}: Instructing retrieval augmented generation via self-synthesized rationales}.
\newblock In \emph{Adaptive Foundation Models: Evolving AI for Personalized and Efficient Learning}.

\bibitem[{Weller et~al.(2024)Weller, Marone, Weir, Lawrie, Khashabi, and Van~Durme}]{pretrainattribution2}
Orion Weller, Marc Marone, Nathaniel Weir, Dawn Lawrie, Daniel Khashabi, and Benjamin Van~Durme. 2024.
\newblock \href {https://aclanthology.org/2024.eacl-long.140} {{``}according to . . . {''}: Prompting language models improves quoting from pre-training data}.
\newblock In \emph{Proceedings of the 18th Conference of the European Chapter of the Association for Computational Linguistics (Volume 1: Long Papers)}, pages 2288--2301, St. Julian{'}s, Malta. Association for Computational Linguistics.

\bibitem[{Wu et~al.(2024)Wu, Xie, Chen, Zhu, Zhang, and Xiao}]{wu2024easilyirrelevantinputsskew}
Siye Wu, Jian Xie, Jiangjie Chen, Tinghui Zhu, Kai Zhang, and Yanghua Xiao. 2024.
\newblock \href {https://arxiv.org/abs/2404.03302} {How easily do irrelevant inputs skew the responses of large language models?}
\newblock \emph{Preprint}, arXiv:2404.03302.

\bibitem[{Xia et~al.(2015)Xia, Xu, Lan, Guo, and Cheng}]{learningmaximalmarginal}
Long Xia, Jun Xu, Yanyan Lan, Jiafeng Guo, and Xueqi Cheng. 2015.
\newblock \href {https://doi.org/10.1145/2766462.2767710} {Learning maximal marginal relevance model via directly optimizing diversity evaluation measures}.
\newblock In \emph{Proceedings of the 38th International ACM SIGIR Conference on Research and Development in Information Retrieval}, SIGIR '15, page 113–122, New York, NY, USA. Association for Computing Machinery.

\bibitem[{Xiao et~al.(2023)Xiao, Liu, Zhang, and Muennighoff}]{bge_embedding}
Shitao Xiao, Zheng Liu, Peitian Zhang, and Niklas Muennighoff. 2023.
\newblock \href {https://arxiv.org/abs/2309.07597} {C-pack: Packaged resources to advance general chinese embedding}.
\newblock \emph{Preprint}, arXiv:2309.07597.

\bibitem[{Xiong et~al.(2020)Xiong, Xiong, Li, Tang, Liu, Bennett, Ahmed, and Overwijk}]{ance}
Lee Xiong, Chenyan Xiong, Ye~Li, Kwok-Fung Tang, Jialin Liu, Paul Bennett, Junaid Ahmed, and Arnold Overwijk. 2020.
\newblock \href {https://arxiv.org/abs/2007.00808} {Approximate nearest neighbor negative contrastive learning for dense text retrieval}.
\newblock \emph{Preprint}, arXiv:2007.00808.

\bibitem[{Xu et~al.(2024)Xu, Gao, Yu, Bi, Shen, and Cheng}]{xu2024aliiceevaluatingpositionalfinegrained}
Yilong Xu, Jinhua Gao, Xiaoming Yu, Baolong Bi, Huawei Shen, and Xueqi Cheng. 2024.
\newblock \href {https://arxiv.org/abs/2406.13375} {Aliice: Evaluating positional fine-grained citation generation}.
\newblock \emph{Preprint}, arXiv:2406.13375.

\bibitem[{Yang et~al.(2018)Yang, Qi, Zhang, Bengio, Cohen, Salakhutdinov, and Manning}]{yang-etal-2018-hotpotqa}
Zhilin Yang, Peng Qi, Saizheng Zhang, Yoshua Bengio, William Cohen, Ruslan Salakhutdinov, and Christopher~D. Manning. 2018.
\newblock \href {https://doi.org/10.18653/v1/D18-1259} {{H}otpot{QA}: A dataset for diverse, explainable multi-hop question answering}.
\newblock In \emph{Proceedings of the 2018 Conference on Empirical Methods in Natural Language Processing}, pages 2369--2380, Brussels, Belgium. Association for Computational Linguistics.

\bibitem[{Yu et~al.(2024)Yu, Ping, Liu, Wang, You, Zhang, Shoeybi, and Catanzaro}]{rankrag}
Yue Yu, Wei Ping, Zihan Liu, Boxin Wang, Jiaxuan You, Chao Zhang, Mohammad Shoeybi, and Bryan Catanzaro. 2024.
\newblock \href {https://arxiv.org/abs/2407.02485} {Rankrag: Unifying context ranking with retrieval-augmented generation in llms}.
\newblock \emph{Preprint}, arXiv:2407.02485.

\bibitem[{Yu et~al.(2023)Yu, Xiong, Yu, and Liu}]{aar}
Zichun Yu, Chenyan Xiong, Shi Yu, and Zhiyuan Liu. 2023.
\newblock \href {https://doi.org/10.18653/v1/2023.acl-long.136} {Augmentation-adapted retriever improves generalization of language models as generic plug-in}.
\newblock In \emph{Proceedings of the 61st Annual Meeting of the Association for Computational Linguistics (Volume 1: Long Papers)}, pages 2421--2436, Toronto, Canada. Association for Computational Linguistics.

\bibitem[{Zamani and Bendersky(2024)}]{stochasticrag}
Hamed Zamani and Michael Bendersky. 2024.
\newblock \href {https://doi.org/10.1145/3626772.3657923} {Stochastic rag: End-to-end retrieval-augmented generation through expected utility maximization}.
\newblock In \emph{Proceedings of the 47th International ACM SIGIR Conference on Research and Development in Information Retrieval}, SIGIR '24, page 2641–2646, New York, NY, USA. Association for Computing Machinery.

\bibitem[{Zhang et~al.(2024)Zhang, Zhang, Guo, de~Rijke, Fan, and Cheng}]{arellmgoodatutilityjudgements}
Hengran Zhang, Ruqing Zhang, Jiafeng Guo, Maarten de~Rijke, Yixing Fan, and Xueqi Cheng. 2024.
\newblock \href {https://doi.org/10.1145/3626772.3657784} {Are large language models good at utility judgments?}
\newblock In \emph{Proceedings of the 47th International ACM SIGIR Conference on Research and Development in Information Retrieval}, SIGIR '24, page 1941–1951, New York, NY, USA. Association for Computing Machinery.

\bibitem[{Zhao et~al.(2024)Zhao, Zhou, Li, Tang, Wang, Hou, Min, Zhang, Zhang, Dong, Du, Yang, Chen, Chen, Jiang, Ren, Li, Tang, Liu, Liu, Nie, and Wen}]{llmsurvey1}
Wayne~Xin Zhao, Kun Zhou, Junyi Li, Tianyi Tang, Xiaolei Wang, Yupeng Hou, Yingqian Min, Beichen Zhang, Junjie Zhang, Zican Dong, Yifan Du, Chen Yang, Yushuo Chen, Zhipeng Chen, Jinhao Jiang, Ruiyang Ren, Yifan Li, Xinyu Tang, Zikang Liu, Peiyu Liu, Jian-Yun Nie, and Ji-Rong Wen. 2024.
\newblock \href {https://arxiv.org/abs/2303.18223} {A survey of large language models}.
\newblock \emph{Preprint}, arXiv:2303.18223.

\end{thebibliography}

\clearpage
\appendix

\section{Details of Data Synthesis}
\label{sec:appendix_a}

The detailed steps of the data synthesis pipeline in SCARLet are described as follows:

\paragraph{Seed Datasets Collection} We first collect seed data for the data synthesis pipeline. A task pool is defined, including the selected tasks and their corresponding datasets. Each task is associated with task instruction and retrieval instruction, as shown in Table \ref{tab:tb-6}. For each dataset, we randomly sample 1,000 instances from its training split, including both input and ground truth. Every sampling uses the same random seed for every dataset.

\paragraph{Entities Extraction} For each seed data instance, we extract entities for subsequent passages retrieval. We utilize the SpaCy\footnote{\url{https://spacy.io/}} toolkit to extract entities from both the input and ground truth. Data instances without extractable entities are discarded.

\paragraph{Entities Retrieval} This stage is to retrieve more relevant entities based on the extracted ones. This serves two purposes: 1) to enhance diversity, and 2) to strengthen relationships between entities, facilitating better construction of the shared context. We retrieve neighboring entities from Wikidata, considering only the direct related entities of each existing entity. To achieve this, we write the SPARQL query for retrieval, as shown below:

\begin{lstlisting}[style=SQLStyle]
SELECT ?property ?propertyLabel ?object ?objectLabel
WHERE {
    wd:{id} ?property ?object.
    ?property rdfs:label ?propertyLabel.
    ?object rdfs:label ?objectLabel.
    FILTER(LANG(?propertyLabel) = "en")
    FILTER(LANG(?objectLabel) = "en")
}
LIMIT {limit}
\end{lstlisting}

\paragraph{Passages Retrieval} After obtaining the expanded entity list, we retrieve relevant passages based on these entities to construct the shared context.

\paragraph{Data Synthesis} At this stage, training data is synthesized for different tasks in the task pool based on the shared context. First, a synthesizer model is selected, which must possess sufficient reasoning and generation capabilities to ensure the quality of the synthetic data. To help the synthesizer understand the task definition and follow the correct format, we provide task instruction, task description, and example data in the prompt. The synthetic data should include both the input and ground truth. The prompt template we use is shown in Table \ref{tab:tb-9}.

\paragraph{Data Filtering} In this stage, the data synthesized in the previous phase is cleaned to further ensure data quality and training stability. We prompt the synthesizer model to check the synthetic data for logical consistency and format correctness based on the shared context. The prompt used for this stage is shown in Table \ref{tab:tb-10}.

\paragraph{Passages Enhancement} To enhance the robustness of the training, we inject noise into the shared context. We instruct the synthesizer model to generate a passage that is semantically relevant but useless for downstream task, and then add this passage to the shared context. The prompt used for this stage is shown in Table \ref{tab:tb-11}.

\begin{table*}[p]
\center
\small
\begin{tabular}{p{0.1\linewidth}  p{0.2\linewidth}  p{0.35\linewidth}  p{0.25\linewidth}}
\toprule
\textbf{Dataset} & \textbf{Task} & \textbf{Task Instruction} & \textbf{Retrieval Instruction} \\

\hline

\multirow{3}{*}{NQ} & \multirow{3}{*}{Single-hop QA} & \multirow{3}{\linewidth}{\texttt{Answer the question based on the given passages.}} & \multirow{3}{\linewidth}{\texttt{Retrieve passages to answer the question.}} \\
& & & \\
& & & \\

\hline

\multirow{5}{*}{HotpotQA} & \multirow{5}{*}{Multi-hop QA} & \multirow{5}{\linewidth}{\texttt{Answer the question based on the given passages. You may need to refer to multiple passages.}} & \multirow{5}{\linewidth}{\texttt{Find passages that provide useful information to answer this question.}} \\
& & & \\
& & & \\
& & & \\
& & & \\

\hline

\multirow{5}{*}{ELI5} & \multirow{5}{*}{Long-form QA} & \multirow{5}{\linewidth}{\texttt{Answer the question based on the given passages. The answer needs to be detailed, paragraph-level, and with explanations.}} & \multirow{5}{\linewidth}{\texttt{Retrieve passages that provide a piece of good evidence for the answer.}} \\
& & & \\
& & & \\
& & & \\
& & & \\

\hline

\multirow{5}{*}{FEVER} & \multirow{5}{*}{Fact Checking} & \multirow{5}{\linewidth}{\texttt{Verify whether the claim is correct based on the given passages. If it is correct, output "SUPPORTS", if it is wrong, output "REFUTES".}} & \multirow{5}{\linewidth}{\texttt{Retrieve passages to verify this claim.}} \\
& & & \\
& & & \\
& & & \\
& & & \\

\hline

\multirow{5}{*}{WoW} & \multirow{5}{*}{Dialogue Generation} & \multirow{5}{\linewidth}{\texttt{Generate an appropriate, reasonable and meaningful response based on previous conversations and the following relevant passages.}} & \multirow{5}{\linewidth}{\texttt{Find passages related to the conversation topic.}} \\
& & & \\
& & & \\
& & & \\
& & & \\

\hline

\multirow{7}{*}{T-REx} & \multirow{7}{*}{Slot Filling} & \multirow{7}{\linewidth}{\texttt{Given an entity and an attribute (or relationship), fill in the specific value of the attribute based on the following passages. The entity and the attribute are separated by "[SEP]".}} & \multirow{7}{\linewidth}{\texttt{Find passages related to the entities.}} \\
& & & \\
& & & \\
& & & \\
& & & \\
& & & \\
& & & \\

\hline

\multirow{5}{*}{SciFact} & \multirow{5}{*}{Fact Checking} & \multirow{5}{\linewidth}{\texttt{Verify whether the claim is correct based on the given passages. If it is correct, output "SUPPORT", if it is wrong, output "CONTRADICT".}} & \multirow{5}{\linewidth}{\texttt{Retrieve passages to verify this claim.}} \\
& & & \\
& & & \\
& & & \\
& & & \\

\hline

\multirow{7}{*}{zs-RE} & \multirow{7}{*}{Relation Extraction} & \multirow{7}{\linewidth}{\texttt{Given an entity and an attribute (or relationship), fill in the specific value of the attribute based on the following passages. The entity and the attribute are separated by "[SEP]".}} & \multirow{7}{\linewidth}{\texttt{Find passages related to the entities.}} \\
& & & \\
& & & \\
& & & \\
& & & \\
& & & \\
& & & \\

\hline

\multirow{3}{*}{FiQA} & \multirow{3}{*}{Financial QA} & \multirow{3}{\linewidth}{\texttt{Answer the question based on the given passages.}} & \multirow{3}{\linewidth}{\texttt{Find passages to answer the question.}} \\
& & & \\
& & & \\

\hline

\multirow{9}{\linewidth}{Climate-FEVER} & \multirow{9}{*}{Fact Checking} & \multirow{9}{\linewidth}{\texttt{Verify whether the claim is correct based on the given passages. If it is correct, output "SUPPORTS", if it is wrong, output "REFUTES", if the information is insufficient, output "NOT\_ENOUGH\_INFO", if can't get a sufficiently confident judgment, output "DISPUTED".}} & \multirow{9}{\linewidth}{\texttt{Retrieve passages to verify this claim.}} \\
& & & \\
& & & \\
& & & \\
& & & \\
& & & \\
& & & \\
& & & \\
& & & \\

\bottomrule
\end{tabular}
\caption{Task instructions and retrieval instructions of the datasets in the task pool.}
\label{tab:tb-6}
\end{table*}

\begin{table}[h]
\centering
\small
\begin{tabular}{>{\raggedright\arraybackslash\tt}p{\linewidth}<{}}
\toprule
Please first provide the answer based on the passages that you have ranked in utility and then write the ranked passages in descending order of utility in answering the question, like "My rank: [i]>[j]>...>[k]".\\\\
Context: \color{blue}\{context\}\\\\
\color{black}Question: \color{blue}\{query\}\\
\bottomrule
\end{tabular}
\caption{The prompt template for LLM-based method.}
\label{tab:tb-7}
\end{table}

\section{Details of Utility Attribution}
\label{sec:appendix_b}

\paragraph{Introduction to Attribution} Attribution is a local-interpretable technique used to provide evidence for the model generation~\citep{li2023attributionsurvey, xu2024aliiceevaluatingpositionalfinegrained}. The data source of attribution can be training data \citep{pretrainattribution1, pretrainattribution2}, whereas in RALMs, the source is often retrieved external passages \citep{ragattribution1, ragattribution2}, which we denote as context attribution. Furthermore, contributive attribution is a form of attribution that quantifies the contribution of each data source unit to the generation process. It assigns an attribution score to each unit, where a higher score indicates a greater contribution. In this study, we propose the SCARLet framework, which employs a perturbation-based attribution method to estimate the utility score of each passage within the shared context. Additionally, we evaluate other attribution methods, including attention-based method, gradient-based method, and LLM-based method.

\paragraph{Perturbation-based Method} This method is described in Section \ref{sec:3-3}. Notably, unlike the classical LIME method, we remove the weight of $\mathbf{v}_i$ in the surrogate model, which measures the the cosine distance from the original text. The reason behind this is that for different perturbation vectors, the weight would exacerbate the unfair evaluation of passage utility, as utility features cannot be directly measured by semantic relevance. For passages that are semantically relevant but essentially useless, the variation they bring would be downweighted, as such passages typically cause greater logit fluctuations due to their lack of utility.

\paragraph{Attention-based Method} This method takes the attention score received by each source unit during inference as the attribution score \citep{mylonas2022attentionmatrixdecisionfaithfulnessbased, attentionmeetsposthocinterpretability}. We construct the attention-based baseline by averaging the attention values of each token within each passage, as shown below:
\begin{equation}
    \bm{\alpha}_{d_i}= \frac{1}{K \cdot \left | d_i \right | } \sum_{t \in d_i} \sum_{i=1}^{K} a_t^{\left ( i \right ) },t \in d_i,
\end{equation}
where $\bm{\alpha}_{d_i}$ represents the utility score for passage $d_i$, $K$ indicates the number of attention heads, and $a_t^{\left ( i \right ) }$ indicates the attention value of the $t$-th token in passage $d_i$ of the $i$-th attention head.

\paragraph{Gradient-based Method} This approach determines the utility scores from the gradient of each token in the source unit during backward propagation\citep{gradientbasedattributionmethods, wang2024gradientbasedfeatureattribution}. Specifically, we employ the Gradient times Input \citep[$G \times I$;][]{gtimesi}, which computes the score of each token by performing the dot product as follows:
\begin{equation}
    f_{G\times I}(t)=e_t \cdot \nabla_{e_t} f_{\mathrm{LM} }\left ( x,D \right )  ,
\end{equation}
where $e_t$ represents the embedding vector of token $t$, and $f_{\mathrm{LM}}$ denotes the function of $\mathrm{LM}$. The utility score of each passage is then obtained by averaging the $G\times I$ scores of each token contained within it.

\paragraph{LLM-based Method} This approach, which can also be referred to as rationale-based method or self-rationalization, is in line with the work of \citet{sohn2024rationaleguidedretrievalaugmentedgeneration, wei2024instructrag}, where the LLM generator simultaneously attributes the utility of passages in the context while performing the task. Although this method is theoretically flawed due to the potential influence of hallucinations from LLMs~\citep{chen2024llmgeneratedmisinformationdetected}, we still believe that it represents one of the future directions of utility attribution. Following \citet{arellmgoodatutilityjudgements}, we instruct the generator to rank the passages from the context in a list-wise setup while generating the answer. The prompt is shown at Table \ref{tab:tb-7}.

\paragraph{GTI Benchmark} This benchmark \citep[Ground-Truth Inclusion;][]{arellmgoodatutilityjudgements} is designed to assess the utility of retrieved passages including three QA datasets: NQ, with 1,868 data; HotpotQA, with 4,407 data; and MSMARCO-QA, with 3,121 data. It manually constructs 10 passages per query, including ground truth (correct passages), counterfactual passages, highly relevant noisy passages, and weakly relevant noisy passages. We evaluate the above methods on this benchmark using LLaMA-3-8B-Instruct as the generator, with the experimental results presented in Table \ref{tab:tb-8}. The results demonstrate that the perturbation-based method outperforms all other baselines by a significant margin, highlighting its considerable advantage as an indicator for utility in RALMs.

\begin{table}
    \centering
    \resizebox{\linewidth}{!}{

\renewcommand\arraystretch{1.1}
\begin{tabular}{lcccccc}
    \toprule
    \multirow{3}{*}{\textbf{Method}} & \multicolumn{2}{c}{\textbf{HotpotQA}} & \multicolumn{2}{c}{\textbf{NQ}} & \multicolumn{2}{c}{\textbf{MSMARCO-QA}} \\
    \cmidrule(lr){2-3} \cmidrule(lr){4-5} \cmidrule(lr){6-7}
    & \textbf{NDCG@1} & \textbf{NDCG@5} & \textbf{NDCG@1} & \textbf{NDCG@5} & \textbf{NDCG@1} & \textbf{NDCG@5} \\
    \midrule
    Att.-based & 31.54 & 27.25 & 29.14 & 25.77 & 29.92 & 22.15 \\
    Grad.-based & 49.90 & 38.83 & 50.58 & 44.56 & 59.09 & 53.35 \\
    LLM-based & 76.34 & 76.84 & 28.35 & 32.16 & 31.88 & 59.97 \\
    Pert.-based & \textbf{93.28} & \textbf{83.04} & \textbf{78.16} & \textbf{84.12} & \textbf{91.65} & \textbf{85.36} \\
    \rowcolor{mygrey}
    \quad w/o G.T. & 92.34 & 81.03 & 77.85 & 80.67 & 91.10 & 83.73 \\
    \bottomrule
    \end{tabular}
}
    \caption{The experimental results comparing various utility attribution methods on the GTI benchmark. Attn., Grad., Pert., and G.T. represent Attention, Gradient, Perturbation and Ground Truth, respectively.}
    \label{tab:tb-8}
\end{table}

\begin{figure*}[htbp]
  \includegraphics[width=\linewidth,scale=1.00]{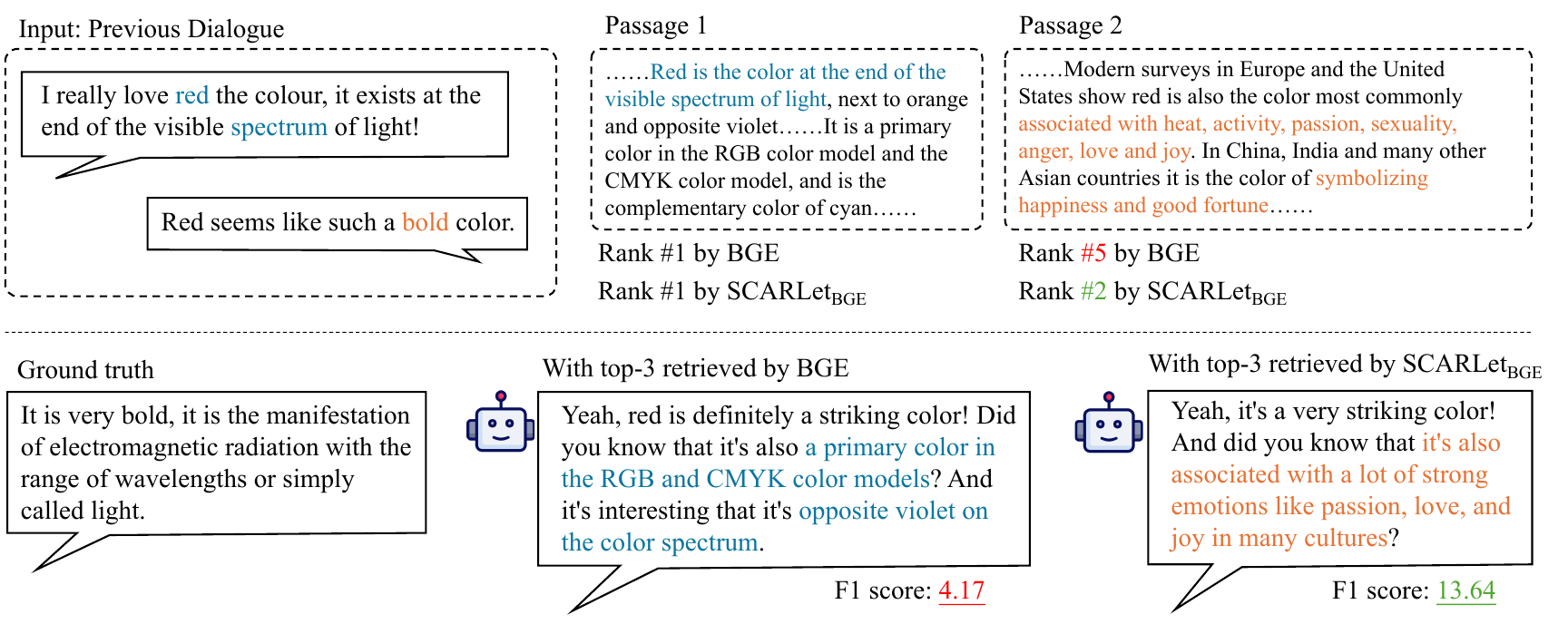}
  \caption {Case Study on WoW. Blue text indicates clues more relevant to semantics, while orange text highlights clues more align with the target utility in dialogue generation task. Responses are generated by LLaMA-3-8B. The generated response augmented by $\textrm{SCARLet}_{\textrm{BGE}}$ achieves a higher F1 score than the response augmented by BGE.}
  \label{fig:fig-6}
\end{figure*}

\paragraph{Attribution Forms} Additionally, we investigate two different attribution forms: 1) The first form directly uses the ground truth provided by the dataset as the output of the generator, which is adopted in our proposed SCARLet; 2) The second form is let the generator to produce a response first, followed by attribution based on that response. The first form reflects the contribution of each passage within the context to the production of the correct answer. While the second form requires an additional comparison between the generated response and the ground truth, where we believe that the attribution process can be valid only if the the two are consistent. We compare the performance of the above two forms in the perturbation-based method, as shown in Table \ref{tab:tb-8}. We find that the performance difference between the two forms is minimal, but in terms of mechanism and difficulty of implementation, we choose the first form.

\section{Details of Implementation}
\label{sec:appendix_c}

\paragraph{Meta data of data synthesis} We present the meta data from the data synthesis pipeline of one run in our experiment, as shown in Table \ref{tab:tb-12}. As observed, although the amount of training data is sufficient for tuning the retriever, the SCARLet pipeline leads to data loss at each stage, sometimes resulting in significant loss rates, which causes an increase in costs. The reasons for the loss include issues with the seed data, network problems, model generation errors, among others.

\paragraph{Hyperparameters} During the data synthesis stage, the temperature of the synthesizer model is set to 0.5. In the utility attribution stage, the number of sampled perturbation vectors $n$ is set to 64, with a perturbation probability of 0.5. During training, we set the learning rate as 6e-5, and epochs as 1. All experiments are conducted on NVIDIA A100 GPUs in torch.float32 precision.

\section{Additional Experimental Results}
\label{sec:appendix_d}

The results presented in Table \ref{tab:tb-2} are under the closed corpus setup, i.e., retrievers search passages only from the corpus of the corresponding dataset. In contrast, the pooled corpus setup refers to merging the corpora of different datasets into a single corpus, where all retrieval is performed with the unified corpus. This setup better simulates real-world retrieval scenarios and enables a fairer evaluation of generalization. The experimental results under the pooled corpus setup are shown in Table \ref{tab:tb-13}. All baselines perform similarly to those in the closed corpus setup, and some outperform them, demonstrating generalization of SCARLet on the unified corpus.

\section{Additional Case Study}
\label{sec:appendix_e}

The QA tasks typically focus more on precise answers, whereas dialogue tasks prioritize the coherence between the generated response and the preceding conversation. These two tasks have distinct retrieval utility, with the latter being more vaguely defined. To analyze whether the retriever trained by SCARLet exhibits a diversified retrieval criteria, we select a case from the test split of the WoW dataset, as shown in Figure \ref{fig:fig-6}. In this case, a retriever relying on semantic relevance may primarily focus on topic words such as "red" and "spectrum". However, for dialogue generation, it is also crucial to consider the intent of the previous speaker. Passage 2 is ranked higher by the retriever trained by SCARLet, because it is directly tied to the deeper meaning of the key clue "bold", making it more helpful in sustaining conversational coherence. At comparable recall levels, SCARLet prioritizes passages that offer greater task-specific utility.

\section{Example of Shared Context}
\label{sec:appendix_f}

\begin{figure*}[htbp]
  \includegraphics[width=\linewidth,scale=1.00]{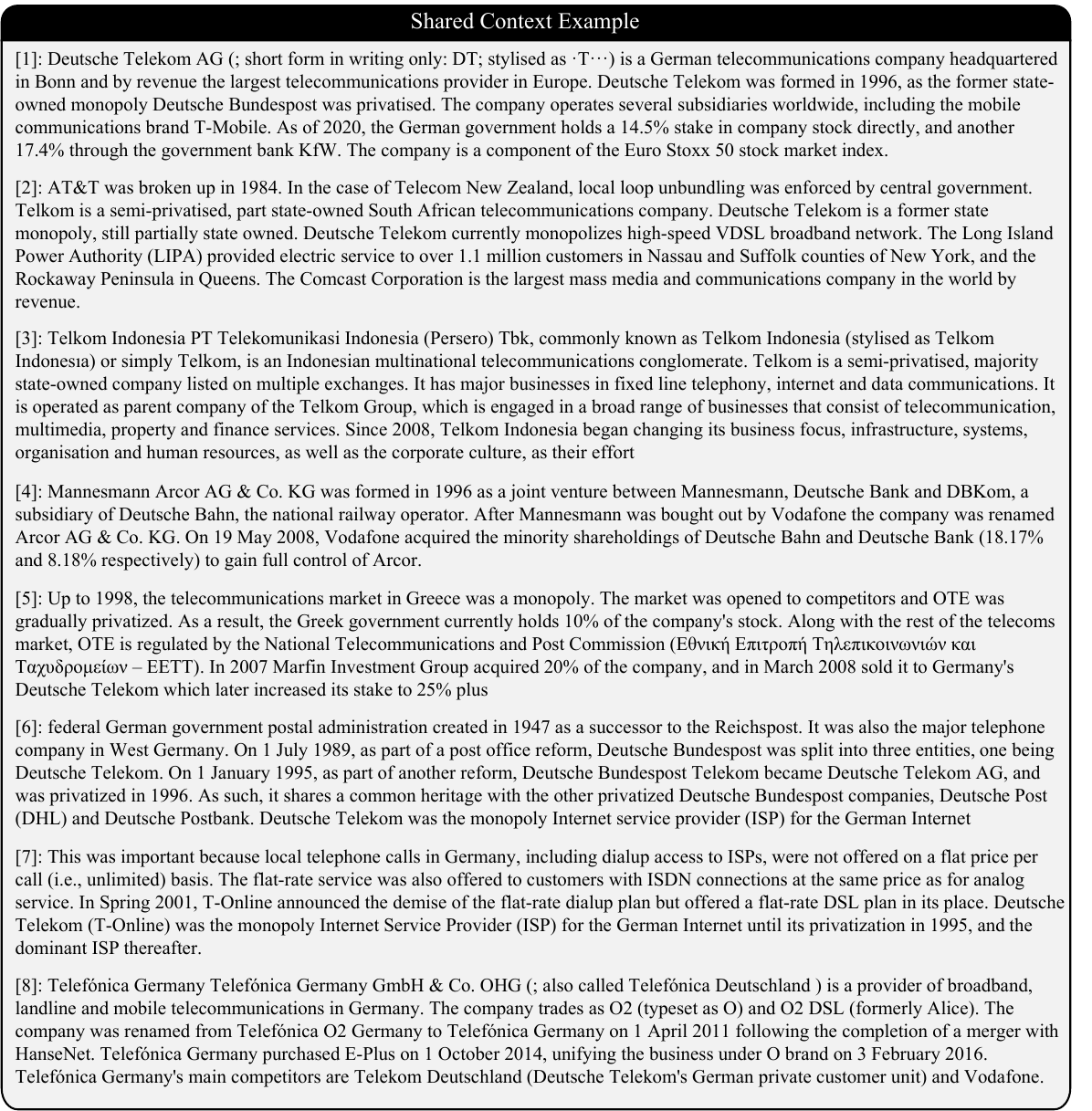}
  \caption {An example of the shared context. Based on this context, SCARLet synthesizes training data, as shown in Figure \ref{fig:fig-8} and \ref{fig:fig-9}.}
  \label{fig:fig-7}
\end{figure*}

\begin{figure*}[htbp]
  \includegraphics[width=\linewidth,scale=1.00]{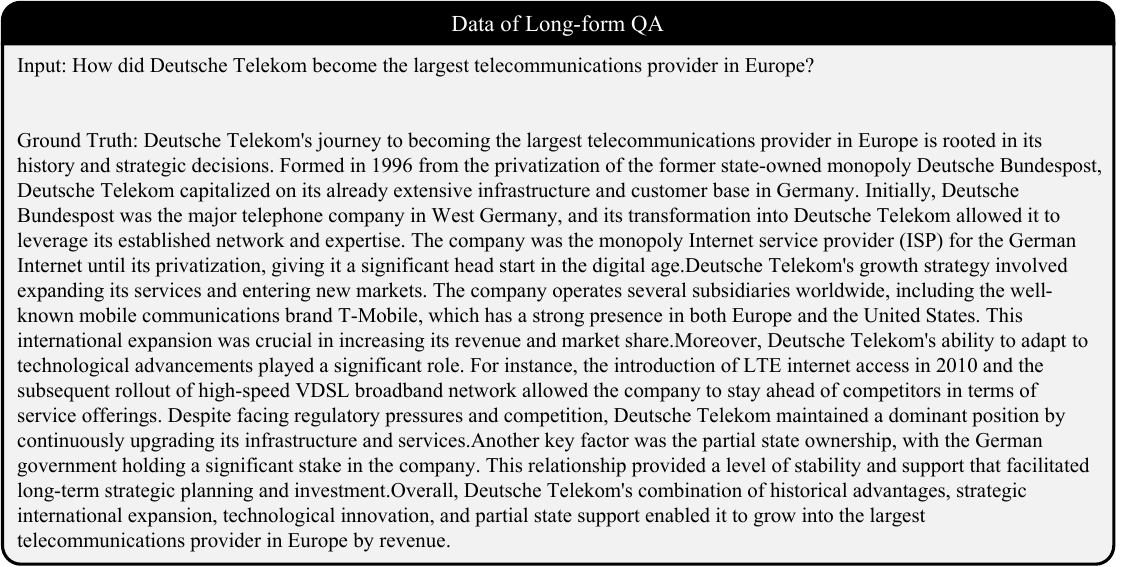}
  \caption {The training data of long-form QA, synthesized by SCARLet based on the context in Figure \ref{fig:fig-7}.}
  \label{fig:fig-8}
\end{figure*}

\begin{figure*}[htbp]
  \includegraphics[width=\linewidth,scale=1.00]{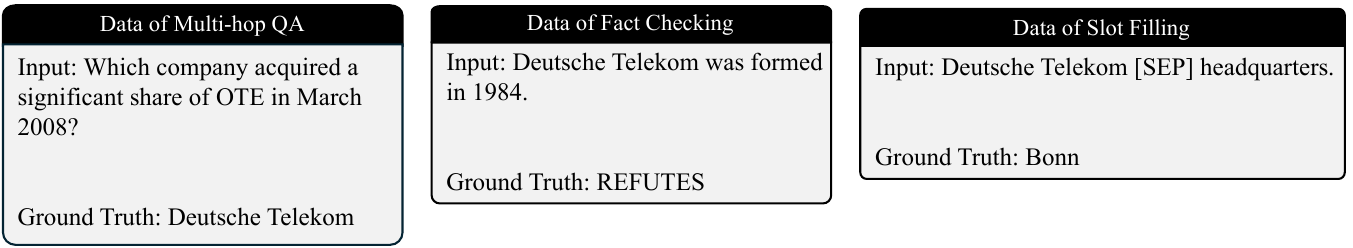}
  \caption {The training data of multi-hop QA, fact checking and slot filling, synthesized by SCARLet based on the context in Figure \ref{fig:fig-7}.}
  \label{fig:fig-9}
\end{figure*}

In this section, we provide an example of the shared context constructed during one run of SCARLet in our experiment, as shown in Figure \ref{fig:fig-7}, along with its corresponding synthetic data for various tasks, as shown in Figure \ref{fig:fig-8} and Figure \ref{fig:fig-9}.

\begin{table*}[ht]
\centering
\small
\begin{tabular}{>{\raggedright\arraybackslash\tt}p{0.9\linewidth}<{}}
\toprule
You are a strong expert of data synthesis. Below, I will provide the context, the description and an example of the target task. Your task is to generate a piece of data for the target task based on the given context. The sections marked with ====xxx begins==== and ====xxx ends==== indicate the start and end of each respective part. Please note that the data you generate must meet the following criteria:\\
1. Correctness, which must be logically correct and factually correct.\\
2. Faithfulness, which must be faithful to the context.\\
3. Quality, which must be thoughtful and sophisticated, ideally based on multiple paragraphs where applicable.\\\\
Please note that the generated data should follow this specific format:\\
====New data begins====\\Input:\\Reference output:\\====New data ends====\\\\
====Context begins====

\color{blue}{\{context\}}\\====Context ends====\\\\
====Target task description begins====

\color{blue}{\{task\_description\}}\\====Target task description ends====\\\\
====Target task example begins====\\Input: \color{blue}{\{task\_example\_input\}}\\Reference output: \color{blue}{\{task\_example\_output\}}\\====Target task example ends====\\\\
Please ensure that your output matches the instructions above.\\

\bottomrule
\end{tabular}
\caption{The prompt template for data synthesis.}
\label{tab:tb-9}
\end{table*}

\begin{table*}[ht]
\centering
\small
\begin{tabular}{>{\raggedright\arraybackslash\tt}p{0.9\linewidth}<{}}
\toprule
You are tasked with checking whether the following synthetic data of \color{blue}{\{task\_name\}} \color{black} task is logically correct and formatted correctly. The data consists of five parts: task description, example, input, output, source passages. The input and output of the synthetic data are based on the source passages. And a reasonable example of \color{blue}{\{task\_name\}} \color{black} task is provided, note that it is not based on source passages. Please check the following:\\
1. Logical Correctness: Check whether the output correctly solves the input based on the source passages.\\
2. Format Correctness: Check whether the input and output of the synthetic data conform to the correct format presented in the task description and the example.\\\\
Task description: \color{blue}{\{task\_description\}}\\\\
Example:\\Input: \color{blue}{\{task\_example\_input\}}\\Output: \color{blue}{\{task\_example\_output\}}\\\\
Now, please check the following synthetic data based on source passages:\\\\
Input: \color{blue}{\{input\}}\\
Output: \color{blue}{\{output\}}\\
Source passages: \color{blue}{\{context\}}\\\\
Please note that if the above synthetic data basically meets the requirements, output "[YES]", otherwise output "[NO]".\\
\bottomrule
\end{tabular}
\caption{The prompt template for data filtering.}
\label{tab:tb-10}
\end{table*}

\begin{table*}[ht]
\centering
\small
\begin{tabular}{>{\raggedright\arraybackslash\tt}p{0.9\linewidth}<{}}
\toprule
You are a strong expert of data processing. You are tasked with data augmentation to generate noisy data to enhance training robustness. Below, I will provide you with a piece of data, including task description, input, and ground truth. Then I will provide you with the context containing the necessary information to solve the input. You need to deeply understand the data and the context, and finally generate a passage which is a variant of one passage of the context. The generated passage needs to be semantically relevant while providing no practical effect in solving the input.\\\\
Data:\\
Input: \color{blue}{\{data\_input\}}\\
Ground truth: \color{blue}{\{data\_output\}}\\\\
Context: \color{blue}{\{context\}}\\\\
Please ensure that the generated passage matches the length of the passages in the context and is a modified version of its original passage. And the generated passage must follow the format, which is marked with ====Generated passage begins==== and ====Generated passage ends==== at its start and end.\\
\bottomrule
\end{tabular}
\caption{The prompt template for passages enhancement.}
\label{tab:tb-11}
\end{table*}

\begin{table*}
\centering
\small
\resizebox{\linewidth}{!}{
\renewcommand\arraystretch{1.1}
\begin{tabular}{lcccccc}
    \toprule
     & \textbf{NQ} & \textbf{HotpotQA} & \textbf{ELI5} & \textbf{FEVER} & \textbf{WoW} & \textbf{T-REx} \\
     \midrule
     \rowcolor{gray!30} \multicolumn{7}{c}{\textbf{Entities Extraction}} \\
     Loss Rate & 12.2\% & 0.6\% & 24.1\% & 5.1\% & 17.1\% & 9.2\% \\
     Averaged Number of Entities & 1.7 & 3.4 & 5.8 & 2.0 & 5.1 & 1.8 \\
     \rowcolor{gray!30} \multicolumn{7}{c}{\textbf{Entities Retrieval}} \\
     Expansion Rate & 91.0\% & 90.5\% & 96.0\% & 89.1\% & 97.5\% & 98.0\% \\
     Averaged Number of New Entities & 5.1 & 6.5 & 17.1 & 3.7 & 14.9 & 5.2 \\
     Averaged Number of Entities & 6.3 & 9.3 & 22.2 & 5.3 & 19.6 & 6.9 \\
     \rowcolor{gray!30} \multicolumn{7}{c}{\textbf{Data Synthesis}} \\
     Number of Synthetic Data & 5230 & 5950 & 4492 & 5580 & 4872 & 5317 \\
     Loss Rate & 12.8\% & 0.8\% & 25.1\% & 7.0\% & 18.8\% & 11.4\% \\
    \bottomrule
\end{tabular}
}
    \caption{Meta data from the synthesis pipeline of one run in our experiment. Loss Rate means the proportion of discarded data caused by the process. Expansion Rate means the proportion of data with new entities added. In this run, the data filtering achieves a loss rate of 44.2\%, and the total amount of data used for utility attribution is 17,529.}
    \label{tab:tb-12}
\end{table*}

\begin{table*}
\centering
\small
\resizebox{\linewidth}{!}{
\renewcommand\arraystretch{1.1}
\begin{tabular}{l|cccccc|cccc}
    \toprule
    \multirow{3}{*}{\textbf{Method}} & \multicolumn{6}{c}{\textbf{In-domain}} & \multicolumn{4}{c}{\textbf{Out-of-domain}} \\
    \cmidrule(lr){2-7} \cmidrule(lr){8-11}
     & \textbf{NQ} & \textbf{HotpotQA} & \textbf{ELI5} & \textbf{FEVER} & \textbf{WoW} & \textbf{T-REx} & \textbf{zs-RE} & \textbf{SciFact} & \textbf{C-FEVER} & \textbf{FiQA} \\
     
     \rowcolor{gray!30} \multicolumn{11}{c}{\textbf{LLaMA-3-8B-Instruct}} \\
     Contriever & 44.0 & 36.7 & 14.5 & 79.2 & 8.6 & 33.8 & 20.9 & 68.1 & 38.0 & 16.5 \\
     BGE & \underline{48.0} & \underline{45.4} & 15.2 & \textbf{85.6} & 8.8 & \textbf{39.6} & 24.1 & \underline{80.2} & \textbf{45.9} & \underline{20.8} \\
     $\textrm{AAR}_{\textrm{Contriever}}$ & 46.2 & 41.8 & 15.0 & 77.8 & 8.2 & 35.1 & \underline{24.2} & 70.3 & \underline{42.6} & 16.7 \\
     $\textsc{RePlug}_{\textrm{Contriever}}$ & 44.5 & 39.7 & 13.8 & \underline{81.3} & 9.2 & 33.7 & 23.6 & 72.9 & 41.0 & 18.8  \\
     \midrule
     \rowcolor{mypurple}
     $\textrm{SCARLet}_{\textrm{Contriever}}$ & 45.1 & 42.0 & \underline{15.9} & 80.6 & \underline{10.4} & 36.4 & 22.2 & 74.7 & 42.0 & 17.7  \\
     \rowcolor{mypurple}
     $\textrm{SCARLet}_{\textrm{BGE}}$ & \textbf{49.8} & \textbf{48.3} & \textbf{16.6} & 81.2 & \textbf{12.7} & \underline{37.0} & \textbf{24.7} & \textbf{81.5} & \textbf{45.9} & \textbf{23.1} \\

     \rowcolor{gray!30} \multicolumn{11}{c}{\textbf{Qwen2.5-3B-Instruct}} \\
     Contriever & 31.9 & 28.5 & 14.2 & 67.1 & 10.5 & 27.1 & 14.0 & \textbf{66.5} & 32.8 & 15.5 \\
     BGE & \textbf{48.5} & \underline{44.0} & 13.7 & \textbf{80.4} & 10.2 & \textbf{34.5} & 18.6 & \underline{65.5} & \textbf{37.1} & 18.6 \\
     $\textrm{AAR}_{\textrm{Contriever}}$ & 34.8 & 30.9 & 13.8 & 66.2 & 10.6 & 28.3 & 15.5 & 63.2 & 32.0 & 16.3 \\
     $\textsc{RePlug}_{\textrm{Contriever}}$ & 34.2 & 35.8 & 14.0 & 71.2 & \textbf{12.8} & 26.8 & 16.9 & 60.6 & 30.9 & \underline{18.7} \\
     \midrule
     \rowcolor{mypurple}
     $\textrm{SCARLet}_{\textrm{Contriever}}$ & 39.3 & 36.0 & \underline{14.4} & 70.0 & 11.9 & 28.2 & \textbf{19.1} & 64.9 & 31.8 & 17.3 \\
     \rowcolor{mypurple}
     $\textrm{SCARLet}_{\textrm{BGE}}$ & \underline{45.1} & \textbf{44.7} & \textbf{15.6} & \underline{74.1} & \underline{12.3} & \underline{30.1} & \underline{18.7} & 64.4 & \underline{36.3} & \textbf{20.5} \\

     \bottomrule
\end{tabular}
}
    \caption{Results of the main experiment in the pooled corpus setup. The unified corpus includes corpora of Wikipedia dump, BeIR-SciFact, BeIR-ClimateFEVER and BeIR-FiQA.}
    \label{tab:tb-13}
\end{table*}

\end{document}